\documentclass[letterpaper]{article} 

\usepackage{times}  
\usepackage{helvet}  
\usepackage{courier}  
\usepackage[hyphens]{url}  
\usepackage{graphicx} 
\usepackage{natbib}  
\usepackage{caption} 
\usepackage{algorithm}
\usepackage{algorithmic}

\usepackage{newfloat}
\usepackage{listings}
\usepackage[utf8]{inputenc} 
\usepackage{booktabs}       
\usepackage{amsfonts}       
\usepackage{nicefrac}       
\usepackage{microtype}      
\usepackage[dvipsnames]{xcolor}         
\usepackage{amsmath,amsthm,amsfonts,amssymb}

\usepackage{glossaries}
\usepackage{siunitx}
\usepackage{graphicx}
\usepackage{tcolorbox}

\usepackage{multicol}
\usepackage{multirow}
\usepackage{pifont}

\usepackage{caption}
\usepackage{subcaption}

\usepackage{enumitem}

\usepackage{adjustbox}
\DeclareMathOperator*{\argmax}{arg\,max}

\newcommand{\expect}{\mathbb{E}}

\newcommand{\bj}{\mathbf{j}}
\newcommand{\bl}{\mathbf{l}}
\newcommand{\bd}{\mathbf{d}}

\newcommand{\ptheta}{\boldsymbol{\theta}}

\newcommand{\pv}{\mathbf{v}}
\newcommand{\pw}{\mathbf{w}}
\newcommand{\px}{\mathbf{x}}
\newcommand{\pu}{\mathbf{u}}
\newcommand{\pz}{\mathbf{z}}
\newcommand{\pb}{\mathbf{b}}

\newcommand{\reals}{\mathbb{R}}
\newcommand{\uspace}{\mathbb{U}}

\newcommand{\polys}[2]{\Pi^{#1}[#2]}

\newcommand{\tf}{\tau}

\DeclareFontFamily{U}{mathx}{\hyphenchar\font45}
\DeclareFontShape{U}{mathx}{m}{n}{
      <5> <6> <7> <8> <9> <10>
      <10.95> <12> <14.4> <17.28> <20.74> <24.88>
      mathx10
      }{}
\DeclareSymbolFont{mathx}{U}{mathx}{m}{n}
\DeclareMathSymbol{\bigtimes}{1}{mathx}{"91}

\theoremstyle{definition}
\newtheorem{definition}{Definition}
\newtheorem{problem}{Problem}

\theoremstyle{plain}
\newtheorem{theorem}{Theorem}

\newtheorem{lemma}{Lemma}

\theoremstyle{plain}
\newtheorem{remark}{Remark}

\newif\ifarxiv

\arxivtrue

\pdfoutput=1

\ifarxiv
    \usepackage{aaai2026}  
    \nocopyright
\else
    \usepackage[submission]{aaai2026}  
\fi

\DeclareCaptionStyle{ruled}{labelfont=normalfont,labelsep=colon,strut=off} 
\floatstyle{ruled}
\newfloat{listing}{tb}{lst}{}
\floatname{listing}{Listing}
\title{
Universal Learning of Stochastic Dynamics for Exact Belief Propagation using Bernstein Normalizing Flows 
}

\ifarxiv
    \author {Peter Amorese and Morteza Lahijanian}
\else
    \author {
        Anonymous Submission
    }
\fi

\usepackage{bibentry}

\begin{document}
\maketitle

\begin{abstract}
Predicting the distribution of future states in a stochastic system, known as \textit{belief propagation}, is fundamental to reasoning under uncertainty. However, nonlinear dynamics often make analytical belief propagation intractable, requiring approximate methods.
When the system model is unknown and must be learned from data, a key question arises: \emph{can we learn a model that (i) universally approximates general nonlinear stochastic dynamics, and (ii) supports analytical belief propagation?}
%
This paper establishes the theoretical foundations for a class of models that satisfy both properties. 
The proposed approach combines the expressiveness of \emph{normalizing flows} for density estimation with the analytical tractability of \emph{Bernstein polynomials}.
Empirical results show the efficacy of our learned model over state-of-the-art data-driven methods for belief propagation, especially for highly non-linear systems with non-additive, non-Gaussian noise.

\end{abstract}

\section{Introduction}

At the heart of intelligent reasoning under uncertainty is the ability to predict future random outcomes. Prediction accuracy has profound implications for the safety and effectiveness of real-world systems. Achieving accurate and informative predictions requires both a representative \emph{model} of the underlying stochastic process and a principled method for \emph{reasoning} with that model. When the model is known but nonlinear (or non-Gaussian), reasoning, specifically predicting the future state distribution, known as \textit{belief propagation}, becomes analytically intractable and typically necessitates approximation. When the underlying dynamics are unknown, machine learning provides a data-driven means to model them. 
However, the choice of learning model should not only aim for expressive power but also support the type of reasoning required. In the case of belief propagation, this leads to a central question: \textit{Can we learn a model that (i) universally approximates general nonlinear stochastic dynamics, and (ii) supports analytical belief propagation?}


This work focuses on the above question and lays the theoretical groundwork for a class of general nonlinear models capable of learning complex stochastic systems while supporting \emph{exact} belief propagation. 
The key insight is that, under a Bayesian (Markov chain) framework, belief propagation reduces to two fundamental operations, \emph{multiplication} and \emph{integration}, over two functions: the \emph{prior} and the \emph{conditional} probability density functions (PDFs). If these functions are represented as polynomials, these operations remain exact, since polynomials are closed under both multiplication and integration. This eliminates the need for approximation during propagation. Building on this observation, we develop a learning framework in which polynomials are trained to soundly represent \emph{true} PDFs, i.e., non-negative functions that integrate to one over their domain. 
Specifically, in our approach, dubbed \textit{Bernstein Normalizing Flows} (BNFs),
the underlying PDFs are modeled using 
Bernstein polynomials, which offer both favorable analytical properties and universal approximation capabilities. 
However,
polynomials alone are ill-suited for modeling PDFs over unbounded state spaces.
To overcome this limitation, we leverage insights from normalizing flows \cite{papamakarios2021normalizing} to enable proper universal density estimation with polynomials.
Hence, 
by embedding Bernstein polynomials in the normalizing flow architecture,
we obtain models that can approximate arbitrary nonlinear stochastic systems while supporting efficient and exact belief propagation.


To the best of our knowledge, the proposed BNF framework is the first class of general nonlinear Markov chain models that simultaneously support universal approximation of stochastic dynamics and exact analytical belief propagation. Beyond the introduction of BNFs, this work makes three key contributions: (i) an explicit-constraint-free training procedure for learning valid PDFs, (ii) a method for enhancing model expressiveness without increasing the number of parameters, and (iii) empirical validation demonstrating the effectiveness of BNFs for belief propagation in comparison with state-of-the-art data-driven approaches.

\subsubsection{Related Work}

Belief propagation in nonlinear stochastic systems is a fundamental challenge in probabilistic reasoning and control. A variety of approximate methods have been developed to handle this task, particularly in the contexts of filtering, density estimation, and probabilistic modeling. These approaches vary in their trade-offs between accuracy, tractability, and the ability to provide formal guarantees.


Approximate belief propagation has been extensively studied in the context of nonlinear filtering \cite{schei1997finite, julier2004unscented}. 
Gaussian Mixture Models (GMMs) are commonly used to represent multimodal beliefs \cite{alspach2003nonlinear} while preserving some analytical tractability, such as efficient sampling and region-based integration, features valuable for planning. However, linearization-based methods like the Extended Kalman Filter (EKF) \cite{jazwinski2013stochastic} introduce significant error under nonlinear dynamics.
Component splitting techniques \cite{kulik2024nonlinearity} mitigate this by recursively subdividing GMM components, but the resulting exponential growth in components renders long-horizon prediction intractable. These filtering approaches also typically assume additive Gaussian noise, making Gaussian Process (GP) regression a common choice for learning system dynamics \cite{deisenroth2011robust}.


Monte Carlo methods \cite{djuric2003particle} offer a flexible alternative by representing beliefs with particle sets. Propagation is straightforward if state transitions can be sampled, but obtaining a usable PDF requires post hoc density estimation. These methods are also prone to inaccuracies in low-sample regions and lack formal error bounds.
In contrast to the approximate methods, our approach performs \emph{exact} belief propagation.
Importantly, similar to GP regression with GMM propagation, our model can evaluate probabilistic future events, and easily draw samples from the current belief.


Several works have aimed to provide \emph{formal} error bounds on belief propagation. For instance, \citet{polymenakos2020safety} derives bounds for GP models, while grid-based techniques generalize formal GMM propagation to broader nonlinear systems \cite{figueiredo2024uncertainty}, albeit under restrictive assumptions (e.g., diagonal-covariance GMM approximations). Moment-based propagation methods also exist \cite{jasour2021moment}, but moments alone are often insufficient to fully recover the underlying distribution \cite{akhiezer2020classical}. 
Unlike these methods, our approach does not assume a given model; instead, we focus on learning a model that supports exact belief propagation, eliminating the need for bounding the approximation error entirely.


Polynomials have long been used for density estimation \cite{yu2023polynomial} due to their normalization convenience.
\ifarxiv
    Bernstein polynomials, in particular, applied to smooth approximations of empirical distributions and in nonparametric settings \cite{babu2002application, belalia2017smooth}, though often via kernel-based methods. Bernstein polynomials also provide easy-to-calculate bounds, making them a powerful tool for constrained functional optimization \cite{amorese2025polynomial}.
\else
    Bernstein polynomials, in particular, have been applied to smooth approximations of empirical distributions and in nonparametric settings \cite{babu2002application, belalia2017smooth}, though often via kernel-based methods. 
\fi
Sampling from general multivariate polynomial densities remains a challenge. 
Our work addresses this by embedding Bernstein polynomials within a normalizing flow architecture. Univariate Bernstein polynomials have been used in normalizing flows in conjunction with neural networks for robustness \cite{ramasinghe2021robust}. However, due to the neural network, the modeled density is not a polynomial.
In contrast, our proposed model is entirely polynomial, enabling efficient sampling, maintaining model sparsity, and supporting exact belief propagation.

\paragraph{Notation} 
Throughout the paper, bold symbols denote vectors, e.g., $\px = [x_1, \ldots, x_n] \in \mathbb{R}^n$, and polynomial functions are denoted as $\pi(\px)$. Given a random variable $\px$, its probability density function (PDF), which we also refer to as \textit{density}, is denoted $p(\px)$, and the probability of $\px$ being in region $R \subseteq \mathbb{R}^n$ is denoted $P(\px \in R)$. 
Multivariate polynomials (e.g., in the monomial-basis) $\pi(\px) = \sum_{j_1=0}^{d_1} \cdots \sum_{j_n=0}^{d_n} c_{j_1, \cdots, j_n} x_1^{j_1} \cdots x_n^{j_n}$ are denoted with multi-indices $\pi(\px) = \sum_{\bj=\mathbf{0}}^{\bd} c_{\bj} \px^{\bj}$, and the set of corresponding coefficients is denoted in bold as $\mathbf{c}$. The set of all degree-$\bd$ polynomials is denoted $\polys{\bd}{\px}$. 
For a vector-valued function $f(\px)$, the $i$-th component is denoted $f_i(\px)$.
If $f$ depends only on the first $l \leq n$ components of $\px$, we denote this by writing $f(\px_{\leq l})$. If $f$ depends only on the $i$-th component of $\px$, we use $f(\px_i)$.
Finally, the \textit{image} of region $R$ via mapping $f$ is denoted $f(R) := \{f(\px) \mid \px \in R \}$.

\section{Problem Formulation} \label{sec:pf}
Consider a discrete-time, non-linear stochastic dynamical system of the form
\begin{equation} \label{eq:system}
    \px_{k+1} = f(\px_k, \pv_k)
\end{equation}
where $\px_k \in X \subseteq \reals^n$
is the state, and $\pv_k \sim p(\pv)$ is a stochastic process noise variable. We assume function $f$ is continuous and differentiable, and
the initial state $\px_0 \sim p(\px_0)$ is random. 
This work considers a state space $X$ that is either unbounded, i.e., $X = \reals^n$, or bounded hyperrectangular, i.e., $X = \bigtimes_{i=1}^n I_i$ for some closed intervals $I_i \subset \mathbb{R}$.

The dynamics of 
System~\eqref{eq:system} can be equivalently represented by the state-transition distribution 
\begin{equation}
    p(\px_{k} \mid \px_{k-1}) = f(\px_{k-1}, \cdot)_\# p(\pv)
\end{equation}
where $\#$ denotes the push-forward operator.
This representation, 
which implicitly captures 
the non-linear dependence of $f$ on $\px_k$ and $\pv_k$,
may be very complex for highly non-linear systems. 
For ease of presentation, we denote this distribution as $p(\px' \mid \px)$.
Hence, the stochastic evolution of the system \eqref{eq:system} from the initial distribution $p(\px_0)$ is captured by a continuous-state Markov chain $\mathcal{M}_f = (X, p(\px_0), p(\px' \mid \px))$.

We consider a scenario where both the dynamics $f$ and the initial state distribution (belief) $p(\px_0)$ are \emph{unknown} and must be learned from data. Specifically, we assume data is provided in the form of state input-output pairs $\mathcal{D}_{\px'} = \{(\hat{\px}, \hat{\px}')_i\}_{i=1}^{N}$, 
where 
$\hat{\px} \in X$ is a point and
$\hat{\px}' = f(\hat{\px}, \pv)$ for a realization of $\pv$,
and initial states $\mathcal{D}_{\px_0} = \{(\hat{\px}_0)_i\}_{i=1}^{N_0}$ for $N,N_0 \in \mathbb{N}$.

We are interested in belief propagation, i.e., representing the (marginal) state distribution $p(\px_K)$, also referred to as the \emph{belief} of the state at some future time $K \in \mathbb{N}$, 
and probabilistic reachability of a region of interest $R \subset X$ at time step $K$, i.e., $P(\px_K \in R)$, 
subject to starting at the initial belief $p(\px_0)$. 
Using the Markov chain $\mathcal{M}_f$,
the belief can be propagated recursively for $k = 1,\ldots, K$, using:
\begin{equation} \label{eq:prop}
    p(\px_{k}) = \int_{\px_{k-1}} p(\px_{k} \mid \px_{k-1}) p(\px_{k-1}) d\px_{k-1}.
\end{equation}
Once $p(\px_{K})$ is constructed, 
then, the probabilistic reachability is given by
\begin{equation} \label{eq:eval}
    P(\px_K \in R) = \int_R p(\px_K) d\px_K.
\end{equation}
We refer to \eqref{eq:eval} as belief \textit{evaluation}.

\subsubsection*{Challenges}
The analytical feasibility of the integral operations described by \eqref{eq:prop} and \eqref{eq:eval} is critically dependent on the coupling between the functional form of both distributions $p(\px_k)$ and $p(\px' \mid \px)$.
In fact, for most known systems that have non-linear dynamics, 
or even linear $f$ but non-Gaussian $p(\pv)$, propagation via \eqref{eq:prop} is analytically infeasible, necessitating approximation.

Since both $p(\px_0)$ and $p(\px' \mid \px)$ are unknown and must be learned, the following question arises: \textit{Is there a functional form for $p(\px_k)$ and $p(\px_{k} \mid \px_{k-1})$ that can model general non-linear stochastic dynamics such that \eqref{eq:prop} and \eqref{eq:eval} become analytically tractable?} 
To rigorously differentiate such desireable models from linear-Gaussian models, or even non-linear models that assume \textit{additive-Gaussian} noise, we define a universal distribution approximator.

\begin{definition}[Universal Distribution Approximator] \label{def:universal}
    Let $\mathcal{P}$ be the set of all continuous PDFs supported on a compact bounded support $\mathcal{X} \subset \reals^n$, i.e., $p(\mathrm{x}) > 0$ for all $p \in \mathcal{P}$, $\mathrm{x} \in \mathcal{X}$. Let $\mathcal{P}_\theta$ be a family of distributions parameterized by a set of parameters $\theta$ where $|\theta| < \infty$. Then, under any probability divergence $\mathcal{D}(\cdot || \cdot)$, $p_\theta \in \mathcal{P}_\theta$ is said to be \emph{universal} if for any $p \in \mathcal{P}$ and $\epsilon > 0$, if
    there exists a $\theta$ such that $\mathcal{D}(p || p_\theta) < \epsilon$.
\end{definition}

\noindent Definition \ref{def:universal} can be easily extended to describe universal \textit{conditional} PDF approximators by asserting that $p_\theta(\mathrm{x} \mid \mathrm{y})$ is universal for all $\mathrm{y}$ in a compact bounded region $\mathcal{Y}$. 



This work tackles the challenge of choosing a functional form for learning an arbitrary Markov chain $\mathcal{M}_f$ subject to:
\begin{enumerate}[label=C\arabic*, labelsep=-1em]
    \item\;\;\;\,. \ $p_\theta(\px_0)$ and $p_\theta(\px_{k} \mid \px_{k-1})$ are \textit{universal},  \label{cond:universal}
    \item\;\;\;\,. \ $p_\theta(\px_k)$ can be computed exactly via \eqref{eq:prop}, and               \label{cond:analytical}
    \item\;\;\;\,. \ $P(\px_k \in R)$ can be computed exactly via \eqref{eq:eval} for any 
    \\
    \textcolor{white}{.}\;\;\;\;\, hyper-rectangle $R \subset X$.               
    \label{cond:evaluation}
\end{enumerate}
Hence, the belief propagation and evaluation problem that we consider is as follows.
\begin{problem}
    \label{problem}
    Given datasets $\mathcal{D}_{\px_0}$ and $\mathcal{D}_{\px'}$ generated from the stochastic System~\eqref{eq:system} and a time horizon $K \in \mathbb{N}$, learn a Markov chain $\mathcal{M}_f$ that adheres to conditions~\ref{cond:universal}-\ref{cond:evaluation}, and compute the distribution $p(\px_K)$.
\end{problem}

\subsubsection{Approach Overview}

Ensuring that all three constraints hold poses a challenging functional representation problem. Recall that the recursive operation in \eqref{eq:prop} involves \emph{multiplication} and \emph{marginalization}, while \eqref{eq:eval} requires \emph{definite integration}. Among common function classes, \emph{polynomials} are one of the few that are closed under these operations and are also known for their strong approximation capabilities. We leverage these properties to address Problem~\ref{problem}.


In the remainder of the paper, all probability densities $p_\theta(\cdot)$ assumed to be parameterized models. For notational simplicity, we omit the subscript $\theta$.  
\ifarxiv
    All proofs are provided in the Appendix.
\else
    All proofs are provided in the Appendix, included in the supplementary material.
\fi
\section{Preliminaries} \label{sec:prelim}

Our approach builds on two existing concepts, normalizing flows and Bernstein polynomials, which we  review here.

\subsubsection{Normalizing Flow} \label{sec:arnf}
Normalizing flows \cite{papamakarios2021normalizing}
are a powerful and expressive parametric form for distribution modeling. In essence, a normalizing flow consists of two parts: i) a diffeomorphism (invertible and differential map) $g : X \rightarrow Z$ which maps from a ``feature'' space $X$ to a  ``latent'' space $Z$ that is homeomorphic to $X$, and ii) a simple distribution $p_\pz(\pz)$ (often the standard Gaussian) over the latent space $Z$. 
In essence, $g^{-1}$ transforms $p_\pz(\pz)$ into a arbitrary density $p_\px(\px)$ in the feature space $X$, and $g$ ``reverses'' the transformation, allowing $p_\px(\px)$ to be expressed
using the differential change in volume, i.e.,
\begin{equation} \label{eq:nf}
    p_\px(\px) = p_\pz(g(\px)) \ | \det( J_{g}(\px) ) |,
\end{equation}
where 
$\det (J_g)$ is the determinant of the Jacobian of $g$. 

For density estimation or generative modeling, $g$ is typically parameterized with a universal \textit{invertible} function approximator. In order to make $g$ invertible \textit{and} have a tractable Jacobian determinant, triangular maps are often used, i.e., $g_i(\px) = g_i(\px_{\leq i})$, simplifying \eqref{eq:nf} to
\begin{equation} \label{eq:nf_triang}
    p_\px(\px) = p_\pz(g(\px)) \ \Big| \prod_{i=1}^n \frac{\partial g_i}{\partial x_i}(\px_{\leq i}) \Big|.
\end{equation}
To ensure that the triangular map is invertible, $g_i$ must be monotonic along the $i$-th dimension, i.e., $\frac{\partial g_i}{\partial x_i}(\px_{\leq i}) > 0$ for all $\px_{\leq i} \in \reals^i$.
Importantly, under mild conditions
\cite{papamakarios2021normalizing}
, \eqref{eq:nf_triang} can approximate any arbitrary feature distribution 
$p_{\px}$
as long as $g_i$ can approximate any monotonic function along $i$.


\subsubsection{Bernstein Polynomials}
Any multivariate polynomial $\pi(\px)$ of degree $\bd$ can be equivalently expressed in the Bernstein basis as
\begin{equation}
    \pi(\px) = \sum_{\bj=\mathbf{0}}^{\bd} b_{\bj} \phi_\bj^\bd(\px),
\end{equation}
where $\phi_\bj^\bd(\px) = \prod_{i=1}^n \binom{d_i}{j_i}x_i^{j_i} (1-x_i)^{d_i - j_i}$ are the Bernstein basis polynomials,
and $b_\bj$ are the coefficients. A coefficient set in the Bernstein basis is denoted with $\mathbf{b}$.
Expressing a polynomial in the Bernstein basis allows one to easily bound the polynomial using its coefficients. We take advantage of this property in order to construct valid polynomial distribution models.


\section{Polynomial Distribution Modeling} \label{sec:bnf}
In this section, we present our modeling framework for $p(\px_0)$ and $p(\px_k | \px_{k-1})$. We aim to express both density functions as multivariate polynomials to preserve the feasibility of \eqref{eq:prop} and satisfy conditions \ref{cond:analytical} and \ref{cond:evaluation}. 
However, polynomials inherently cannot represent PDFs supported over unbounded domains such as $\reals^n$, since a valid PDF must be a non-negative function that integrates to one over its domain, whereas any non-negative polynomial
$\pi(\px)$ yields $\int_{\reals^n} \pi(\px) d\px = \infty$.
To address this, we first map the unbounded support $\mathbb{R}^n$ to a bounded domain and then define the polynomial PDF over this transformed space.

\subsubsection{Transforming the State Space}
Let $\uspace^n:=(0, 1)^n$ be the $n$-dimensional open unit box.
To enable the use of polynomial density functions, we employ a mapping $\Omega : X \rightarrow \uspace^n$
to transform the original state space $X$ into the unit box $\uspace^n$. As long as $\Omega$ 
is a diffeomorphism and diagonal, i.e., $\Omega_i(\px) = \Omega_i(\px_i)$, densities over $X$ can be equivalently expressed over $\uspace^n$ using 
\eqref{eq:nf} as follows.
\begin{subequations}
    \begin{align}
        p_\px(\px) = & \ p_\pu(\Omega(\px)) \ |\det J_{\Omega}(\px)| \label{eq:changevar} \\
        = & \ p_\pu(\Omega(\px)) \ \prod_{i=1}^n \Big(\frac{d\Omega_i}{d x_i}(x_i) \Big). \label{eq:diagonal}
    \end{align}
\end{subequations}

In fact, a valid choice for each component $\Omega_i$ of $\Omega$ is a univariate cumulative distribution function (CDF) of a continuous distribution supported on $\reals$, e.g., the Gaussian CDF, since the CDFs are monotonic and continuous, i.e., they are diffeomorphisms.
%
%
The properties of the models described hereafter are do not depend on the specific choice of $\Omega$; however, as discussed in the Appendix, certain choices of $\Omega$ may enhance the model's ability to learn accurate representations of the underlying Markov chain.

The properties of $\Omega$ preserve the integration property 
\begin{equation}
    \int_{I_i} p_\px(\px) dx_i = \int_{\Omega(I_i)} p_\pu(\pu) du_i,
\end{equation}
over any interval $I_i$ of an axis of $X$. By extension, marginalizing or integrating $p_\px(\px)$ over hyper-rectangular regions $R \subseteq X$ is equivalent to marginalizing or integrating over its image $\Omega(R)$ in $\uspace^n$. Note that $\Omega(R)$
is guaranteed to also be hyper-rectangular since $\Omega$ is diagonal and monotone. 

This allows us to express a polynomial density $p_\pu(\pu) = \pi(\pu)$, such that marginalizing over $u_i$ 
reduces to simply integrating the polynomial $\pi(\pu)$
over the interval $[0, 1]$. Additionally, it is easy to verify that the product of any two $\uspace^n$-polynomial densities of different random variables, e.g., $p(\pu_k | \pu_{k-1})p(\pu_{k-1})$, as in \eqref{eq:prop}, results in a polynomial and remains integrable. For the remainder of the paper, we drop the subscript when denoting densities, e.g., $p_\pu(\pu) = p(\pu)$.

Since both \textit{multiplication} and \textit{marginalization} operations in \eqref{eq:prop} (including integration in \eqref{eq:eval}) become simple polynomial operations of the equivalent $\uspace^n$-polynomial densities, we can learn the Markov chain $\mathcal{M}_f$ directly in $\uspace^n$ without imposing any restrictions on the generality of the underlying systems or state distributions. Therefore, all that remains is to specify the polynomial model parametric form for learning $p(\pu_{0})$ and $p(\pu_k | \pu_{k-1})$. 

\subsubsection{Polynomial Density Estimation} \label{sec:poly_den_est}
Density estimation is a functional optimization problem with two constraints over the support: i) the integral of the function over the support must be $1$, and ii) the function must be non-negative. 
Regarding (i), the class of all polynomials of a fixed degree $\bd$ that sum to $1$ over $\uspace^n$, can be easily characterized as a linear constraint over coefficients \cite{yu2023polynomial}.

Enforcing constraint (ii), however, is more challenging since even checking if a polynomial is non-negative over $\uspace^n$ is known to be \textbf{NP-hard} \cite{murty1985some}. 
To address this, we employ the Bernstein polynomial basis, which provides a \textit{relaxation}, i.e., a sufficient (but not necessary) condition for bounding a polynomial (and its derivatives) over $\uspace^n$.
We embed this relaxation within a normalizing flow architecture to ensure the necessary invertibility conditions for synthesizing valid polynomial densities.

\subsection{Bernstein Normalizing Flow} \label{sec:poly_den_est}
We propose the Bernstein Normalizing Flow (BNF) for density estimation, particularly for learning $p(\pu_0)$.
Recall, triangular-map normalizing flows are an expressive parametric form for universal density estimation, which can be easily extended to universal \textit{conditional} density estimation.

We aim to formulate a polynomial distribution over $\uspace^n$ via the diffeomorphism $g : \uspace^n \rightarrow \uspace^n$ where the latent space is also the unit box.
In order to ensure that $g$ is a diffeomorphism, it suffices to show that $g_i$ is monotonic in $u_i$ and spans the full range, i.e., 
for all $\pu = (u_1,\ldots,u_n) \in \uspace^n$,
$g_i(\pu) = 0$ if $u_{i}=0$, and $g_i(\pu) = 1$ if $u_{i}=1$.

Suppose each component $g_i$ is a multivariate Bernstein polynomial $g_i(\pu) = \pi_g^i(\pu_{\leq i})$ of degree $\bd$. By making the latent density 
uniform over $\uspace^n$, the normalizing flow density \eqref{eq:nf} simplifies to
\begin{equation} \label{eq:u_nf}
    p(\pu) = \prod_{i=1}^n \frac{\partial \pi_g^i}{\partial u_i}(\pu_{\leq i}),
\end{equation}
which is simply a product of multivariate polynomials.

Hence, the two invertibility conditions of $g$, i.e., monotonicity and coverage of the full range,
can be expressed in differential form as
\begin{subequations}
    \label{eq:inv}
    \begin{align}
        \frac{\partial g_i}{\partial u_i}(\pu_{\leq i}) &> 0 \label{eq:inv_1} \\
        \int_0^1 \frac{\partial g_i}{\partial u_i}(\pu_{\leq i})du_i &= 1  \; \label{eq:inv_2}
    \end{align}
\end{subequations}
for all $\pu_{\leq i} \in \uspace^{i}$. To enforce these conditions on the Bernstein polynomial $\pi_g^i(\pu_{\leq i})$, we can directly parameterize the partial derivative 
$\pi_{\partial g}^i(\pu_{\leq i}) = (\partial \pi_g^i / \partial u_i)(\pu_{\leq i})$
and enforce constraints 
\eqref{eq:inv}
on
the coefficients via the following lemmas:

\begin{lemma}[Bernstein Relaxation \cite{lorentz2012bernstein}] \label{lem:relax}
    A Bernstein polynomial $\pi(\pu)$ with coefficients $\pb_\bj$ is bounded on $\uspace^n$ by the extrema of its coefficients, i.e., $\pi(\pu) > \min_\bj \pb_\bj$ and $\pi(\pu) <  \max_\bj \pb_\bj$ for all $\pu \in \uspace^n$. 
\end{lemma}

\begin{lemma} \label{lem:normalize}
    Let $\pi(\pu)$ be a Bernstein polynomial of degree $\bd = (d_1, \ldots, d_n)$ with coefficients $\pb_\bj = (b_{j_1}, \ldots, b_{jn})$. Then, 
    \begin{equation}
        \int_0^1 \pi(\pu) d u_i = 1 \;\; \iff  \;\; \sum_{j_i = 0}^{d_i} b_{j_i} = d_i + 1 \;\;\; \forall j_i.
    \end{equation}
\end{lemma}

Let $\tilde{\pb}^i$ be the coefficients of $\pi_{\partial g}^i$, 
which has degree $\tilde{\bd}^i = (d_1, \ldots, d_i - 1, \ldots d_n)$. Then, condition \eqref{eq:inv_1} can be easily enforced by constraining $\tilde{\pb}^i_\bj \geq 0$ for all $\mathbf{0} \leq \bj \leq \tilde{\bd}^i$, and condition \eqref{eq:inv_2} can be enforced by ensuring that all $\tilde{\pb}^i_\bj$ sum to $d_i$ along the $i$-th axis. With both constraints, polynomials $\pi^i_g$ correctly parameterize a diffeomorphism, and therefore $p(\pu)$ is a valid density.

\begin{remark} \label{rem:composition}
    To achieve more expressivity, multiple diffeomorphism ``layers'' $g$ can be composed. Doing so, however, raises the degree multiplicatively of the fully-expanded polynomial (which is necessary for belief propagation and evaluation).
\end{remark}

\subsection{Conditional Density Estimation} \label{sec:cond_den_est}
To parameterize a conditional distribution $p(\pu \mid \pw)$ for $\pw \in \uspace^n$, we can construct a flow function $h : \uspace^n \times \uspace^n \rightarrow \uspace^n$, such that given any $\pw$, $h(\pu, \pw)$ is a diffeomorphism in $\pu$.
The differential constraints \eqref{eq:inv_1} and \eqref{eq:inv_2} can be modified accordingly by adding $\pw$ as a parameter.
Making each component $h_i$ of $h$ a polynomial $h_i(\pu, \pw) = \pi^i_h(\pu_{\leq i}, \pw)$ of degree $\mathbf{D} = (d_1, \ldots, d_n, d_1, \ldots, d_n)$, yields
a polynomial \textit{conditional} distribution of the form
\begin{equation} \label{eq:u_cnf}
    p(\pu \mid \pw) = \prod_{i=1}^n \frac{\partial \pi_h^i}{\partial u_i}(\pu_{\leq i}, \pw).
\end{equation}
By ensuring that the coefficients of each $\pi_h^i$ are all non-negative, and sum to $d_i - 1$ along the $i$-th axes, \eqref{eq:u_cnf} is a valid conditional distribution. 

\begin{theorem} \label{thm:universal}
    The conditional Bernstein normalizing flow is a universal conditional distribution approximator.
\end{theorem}

\subsection{Belief Propagation}
By modeling $p(\pu_0)$ as a BNF and $p(\pu' \mid \pu)$ as a conditional BNF, the operations in \eqref{eq:prop} and \eqref{eq:eval} can be carried out exactly using tensor operations on the coefficients of each model. Additionally, both multiplication and integration operations can be carried out entirely within the Bernstein basis, improving the numerical stability of each prediction \cite{farouki1987numerical}. Since belief propagation is exact with respect to the model, all prediction error can be attributed to modeling/learning error (and floating-point numerical inaccuracies).

The following theorem states another crucial consequence of using BNF to model $\mathcal{M}_f$ for belief propagation.
\begin{theorem} \label{thm:future_class}
    Given a state-transition conditional-BNF $p(\pu' \mid \pu)$ that is of
    degree $\bd'$ in $\pu'$, starting from any valid polynomial belief $p(\pu_0)$, $p(\pu_k) \in \polys{\bd'}{\pu}$ for every $k \in \mathbb{N}$. 
\end{theorem}

Theorem \ref{thm:future_class} has computational implications for long-horizon prediction. Namely, in each propagation recursion \eqref{eq:prop}, the amount of memory needed to store each belief $p(\pu_k)$ does not grow in time, unlike methods such as GMM-splitting \cite{kulik2024nonlinearity}. Additionally, the computation needed for $p(\pu_k)$ is $O(k)$.

We note that the proposed BNF methods are not limited to just belief propagation and can be trivially extended to other applications as discussed in the following remark.

\begin{remark}[Bayesian Belief Update Analog]
    The proposed methods can be employed for exact Bayesian belief \textit{updating} with learned likelihood and prior models, i.e.,
    $p(\alpha \mid \beta) = p_\theta(\beta \mid \alpha) p_\theta(\alpha) / p(\beta).$
    Computing the normalization constant $p(\beta)$ needed to calculate the posterior belief $p(\alpha \mid \beta)$ often results in an infeasible integral for complex likelihood models. Nonetheless, the multiplication-normalization operation sequence closely parallels the propagation operation described in \eqref{eq:prop}, and thus the models presented in this paper can be readily applied to exact Bayesian Belief updating.
\end{remark}
\begin{remark}
    Using the normalizing flow structure allows one to easily sample from both of the learned distributions. One can easily sample a latent state $\hat{\pz}$ uniformly over $\uspace^n$, then map the sample through the inverse of $g$ or $h$, i.e., $\hat{\pu} = g^{-1}(\hat{\pz})$ or $\hat{\pu} = g^{-1}(\pz; \pw)$.
\end{remark}


\section{Training Procedure} \label{sec:train}
In this section, we outline the procedure for training each piece of the Markov chain. Specifically, we show how the Bernstein relaxation constraints described in Sec. \ref{sec:bnf} can be enforced during training. 
Additionally, we describe a procedure that can be implemented during training to tighten the relaxations and increase the expressivity of the the model without increasing the number of parameters.

\subsection{Constrained Log-Likelihood Optimization}
Maximum Likelihood Estimation (MLE) is a well established objective for density estimation \cite{pan2002maximum}. We aim to maximize the likelihood BNF given state space data $D_{\px'}$ and $D_{\px_0}$. MLE of a $\px$-density with $\px$-data is equivalent to MLE of a $\pu$-density against the data mapped to $\uspace^n$ 
via $\Omega$. Therefore, we can formulate the constrained MLE optimization problem over the parameters of each BNF model. Formally, we can state the optimization problem for learning $p(\pu_0)$ as:
\begin{subequations} \label{eq:opt_init}
\begin{align}
    \argmax_{\tilde{\pb} \in \tilde{\mathbf{B}}} & \ \expect_{\pu_0 \sim p^\star(\pu_0)} \big[ \log p(\pu_0) \big] \label{eq:opt_obj}\\
     \text{subject to}\quad \tilde{\pb}^i &\geq 0  \label{eq:opt_const_1} \\
     \quad \sum_{j_i = 0}^{d_i-1} \tilde{\pb}^i &= d_i \quad \; \forall 1\leq i \leq n \label{eq:opt_const_2} 
\end{align}
\end{subequations}
where $p^\star(\cdot)$ denotes the true distribution, and $\tilde{\mathbf{B}} \ni \pb$ denotes the parameter space. Expectations with respect to $p^\star$ are empirically evaluated over the data in $\mathcal{D}_{\px_0}$. A similar optimization problem can be formulated for training the transition distribution by replacing the objective \eqref{eq:opt_obj} with 
\begin{equation}
    \argmax_{\tilde{\pb}^i} \ \expect_{\pu', \pu \sim p^\star(\pu', \pu)}\big[ \log p(\pu' | \pu) \big], 
\end{equation}
where $p^\star(\pu', \pu)$ denotes the true \textit{joint} distribution that generated $\mathcal{D}_{\px'}$. 
Note that, even though the constraints 
\eqref{eq:opt_const_1} and \eqref{eq:opt_const_2} are \textit{linear}, the objective \eqref{eq:opt_obj} is non-linear and non-convex. Thus, we leverage stochastic gradient descent (SGD) for optimization. 

In order to avoid explicitly constraining the parameter space, we can define a differentiable function $\Psi : \Theta \rightarrow \tilde{\mathbf{B}}_{\text{feas}}$ where $\Theta$ represents an ``unconstrained'' parameter space, and $\tilde{\mathbf{B}}_{\text{feas}} \subset \tilde{\mathbf{B}}$ is the set of feasible parameters, i.e., the set of all $\tilde{\pb}$ that satisfy \eqref{eq:opt_const_1} and \eqref{eq:opt_const_2}. With $\Psi$, SGD can be performed in $\Theta$ \textit{instead} of $\mathbf{B}$, thereby enforcing the constraints implicitly. We can define $\Psi$ for a given coefficient tensor $\pb^i$ as the composition of two operations: 1) map each vector to be positive, and 2) normalize the coefficients to sum to $d_i$ along dimension $i$. Formally, we can construct $\Psi$ as
$\tilde{\pb}^i = \Psi_i(\ptheta^i) = (\sigma \circ \delta) (\ptheta^i),$
where 
$\delta$ is a positive-range element-wise function (e.g., softplus) and $\sigma$ is the normalizing function 
\begin{equation} \label{eq:normalize}
    \sigma(\ptheta^i) = \ptheta_i / \sum_{j_i=0}^{d_i-1} \theta^i_\bj.
\end{equation}
Each operation is differentiable; thus, models parameterized by $\ptheta_0, \ldots, \ptheta_{d^\tf-1}$ can be trained with standard SGD are guaranteed to satisfy the constraints.

\subsection{Tightening the Relaxation}
We have thus far ignored an important aspect inherent to the BNF that can significantly affect the models' expressiveness: the tightness of the Bernstein relaxation in Lemma \ref{lem:relax}. Constraining coefficients $\pb > 0$ is only a \textit{sufficient} (but not necessary) condition for $\pi_{\partial g}^i \geq 0$. Thus $\tilde{\mathbf{B}}_{\text{feas}}$ inner-approximates the set of all degree-$\bd$ polynomial diffeomorphisms, and limits the expressiveness of the model. Increasing the degree of each $\pi^i_g$ improves the expressiveness, at the cost of adding many more parameters to the model. We propose an alternative method to tighten the relaxation during the training procedure \textit{without} adding any parameters to the model.

The proposed procedure lifts a given Bernstein polynomial to a higher degree to reason over feasibility.
Any Bernstein polynomial of degree $\bd$ can be equivalently expressed in a Bernstein basis of a higher degree $\bd^+ \geq \bd$ \cite{lorentz2012bernstein}. 
For simplicity of presentation, we describe the following assuming a one-dimensional Bernstein polynomial; however, all operations readily extend to the multivariate case. A Bernstein basis polynomial $\phi_j^d(x)$ can be written as a linear combination of higher-degree basis polynomials:
\begin{equation} \label{eq:deg_raise}
    \phi_j^d(x) = \frac{d+1-j}{d+1}\phi_j^{d+1}(x) + \frac{j+1}{d+1}\phi_{j+1}^{d+1}(x).
\end{equation}
One can apply \eqref{eq:deg_raise} recursively $d^+-d$ times to build a linear transformation of the original coefficients, in the form $\pb^+ = \mathbf{M}_d^{d^+} \pb$ where $\pb^+$ is the degree-$d^+$ coefficient vector of the same polynomial represented by $\pb$, and $\mathbf{M}_d^{d^+}$ is a $(d^++1)\times(d+1)$ tall matrix. 

\begin{theorem}[\cite{garloff1985convergent}] \label{thm:incr}
    Given a Bernstein polynomial $\pi(\pu)$ of degree $\bd$, let $\underline{\pi} = \inf_{\pu \in \uspace^n} \pi(\pu)$ and $\underline{b} = \min_\bj \pb$. Then, for $\bd^+ \geq \bd$,  
    \begin{equation}
        \underline{b} \leq \underline{\pi} \leq \underline{b} + O(\bd^{-1})
    \end{equation}
    and $\underline{b}$ converges to $\underline{\pi}$ monotonically.
\end{theorem}

\noindent
Theorem \ref{thm:incr} illustrates the convergence of the Bernstein relaxation as the raised-degree is increased.

Constraining degree-$d$ polynomials by the degree $d^+$ representation inherently captures \textit{more} non-negative degree-$d$ polynomials, and as $d^+\rightarrow \infty$, the raised degree constraint captures all non-negative polynomials. In other words, there may exist a non-negative degree-$d$ polynomial with coefficients $\pb$ that minimizes \eqref{eq:opt_obj}, but violates constraint \eqref{eq:opt_const_1} due to the conservatism of the relaxation.

Employing degree-raising during training is not as straightforward as using implicit constraints, as discussed in the previous section. To illustrate the challenge, consider the following sequence of operations within a single SGD iteration:
\begin{enumerate}
    \item Raise the degree of the current (unconstrained) parameters via $\pb^+ = \mathbf{M}_d^{d^+} \pb$.
    \item Replace all negative entries with zero.
    \item Project the rectified parameters back down to the degree-$d$ representation.
    \item Normalize the parameters using \eqref{eq:normalize}.
\end{enumerate}
Unfortunately, even though the rectified $\pb^+ \geq 0$ after step 2, the projection does not necessarily preserve this property.

To address this issue, we train the model with a soft-constraint violation penalty. Then, using an iterative projection, the parameters are moved into the feasible region. The constraint violation penalty loss can be formulated as $\sum_\bj \max(0, \pb^+)$, which is easily calculated with matrix multiplication. Then, to apply a hard constraint to the resulting parameters $\pb$, we iteratively repeat the projection steps 1-3 until $\pb^+ \geq 0$. This training procedure avoids cumbersome feasible-set projections during optimization, while still yielding a feasible solution.

\section{Evaluations} \label{sec:exp}
This section evaluates the efficacy of learned BNF Markov chain models for uncertainty propagation and the effect of the choice of degree on the models' expressiveness. Full details on system parameters, experimental setup, and code as well as addition visualizations are provided in the Appendix and supplementary material.

\paragraph{Comparison Methods} 

We compare BNF to three uncertainty propagation methods based on GMM belief representations: (i) first-order linearization around component means (EKF-style) \cite{jazwinski2013stochastic}, (ii) Whitened Spherical Average Second-Order Stretching (WSASOS) \cite{kulik2024nonlinearity}, and (iii) a grid-based approach from \cite{figueiredo2024uncertainty}. EKF and WSASOS assume nonlinear dynamics with additive Gaussian noise. The grid-based method assumes a diagonal-covariance GMM for $p(\px' | \px)$, but lacks a procedure for constructing a formal GMM approximation from general nonlinear models, so we evaluate it assuming a single-component GMM.


All three methods rely on Gaussian assumptions, so we used GP regression to learn $p(\px' \mid \px)$ and expectation minimization on a GMM to learn $p(\px_0)$. EKF serves as a widely used baseline, WSASOS as a state-of-the-art component-splitting method, and the grid-based approach as a formal method with guaranteed error bounds.

\newcommand{\cropfigsole}[1]{
    \includegraphics[width=\textwidth, trim=55 37 45 20, clip]{#1}%
}

\begin{figure*}[t]
    \centering
    \begin{subfigure}{0.12\textwidth}
        \centering
        \cropfigsole{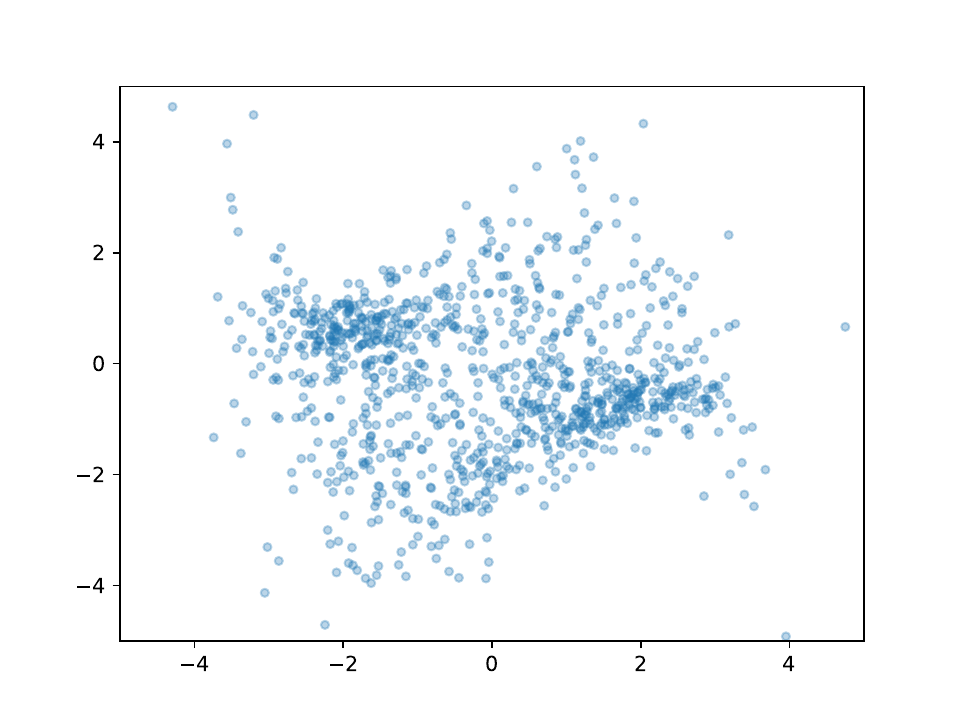} 
        \caption{Monte Carlo}
    \end{subfigure}
    \begin{subfigure}{0.12\textwidth}
        \centering
        \cropfigsole{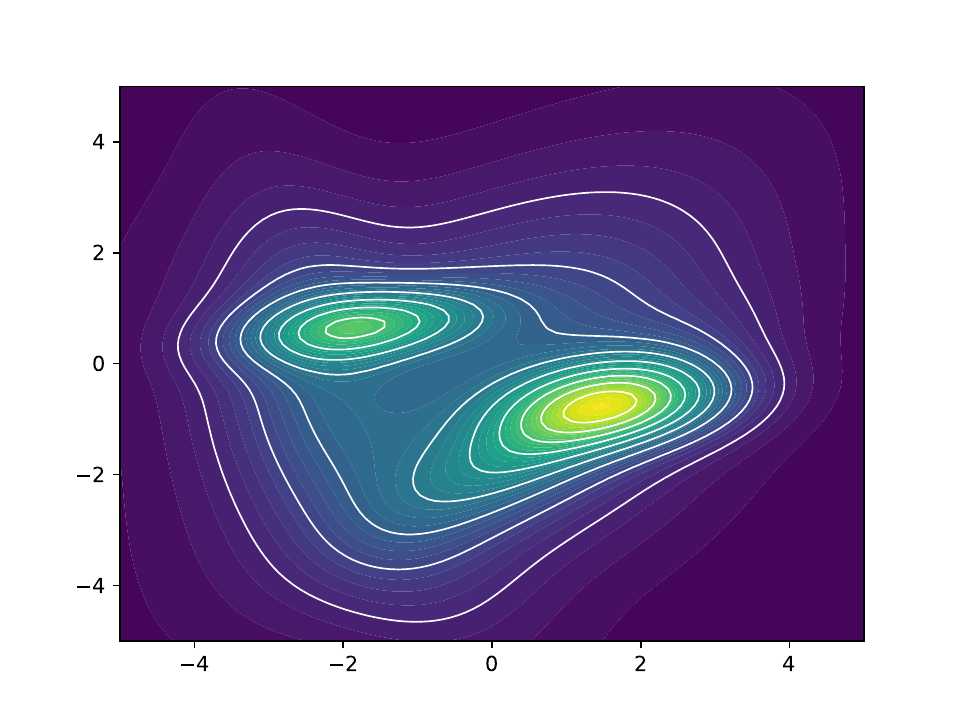}
        \caption{EKF}
    \end{subfigure}
    \begin{subfigure}{0.12\textwidth}
        \centering
        \cropfigsole{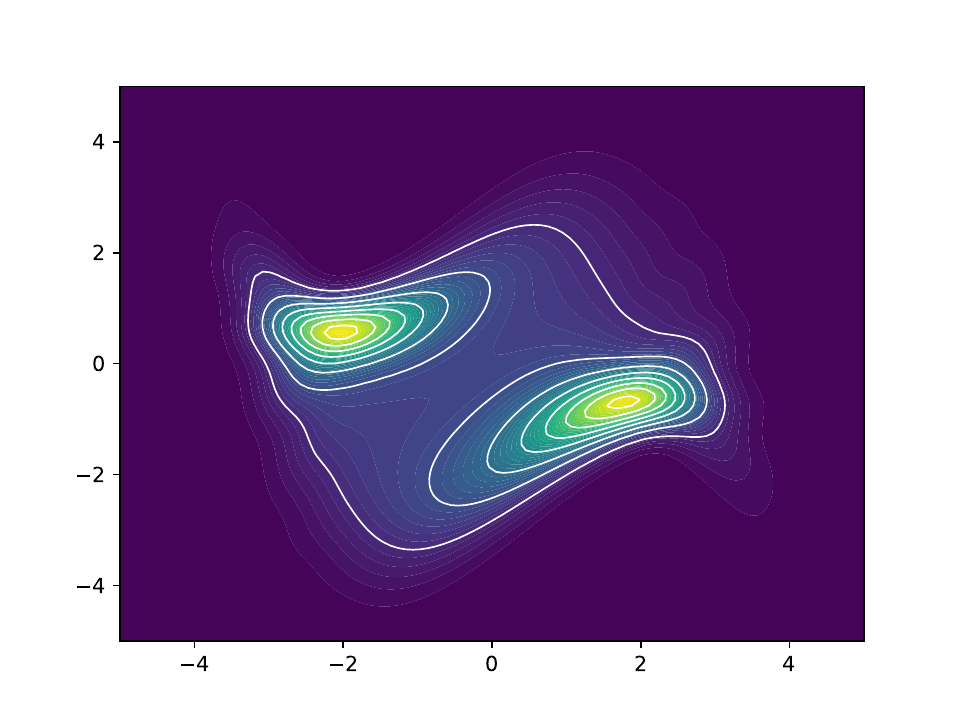}
        \caption{GridGMM}
    \end{subfigure}
    \begin{subfigure}{0.12\textwidth}
        \centering
        \cropfigsole{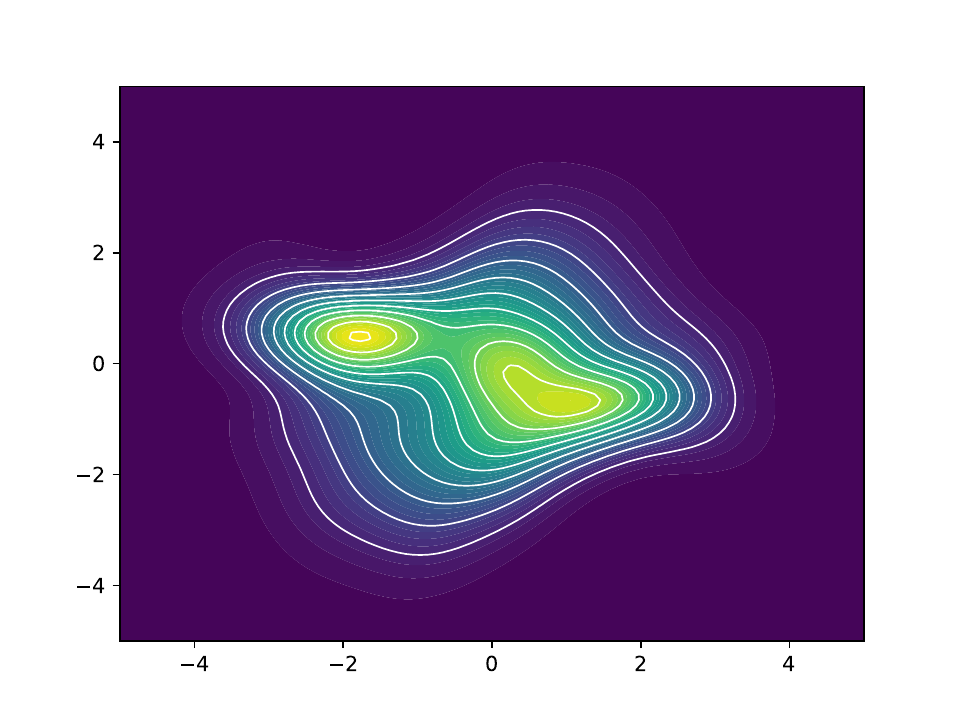}
        \caption{BNF}
    \end{subfigure}
    \
    \begin{subfigure}{0.12\textwidth}
        \centering
        \cropfigsole{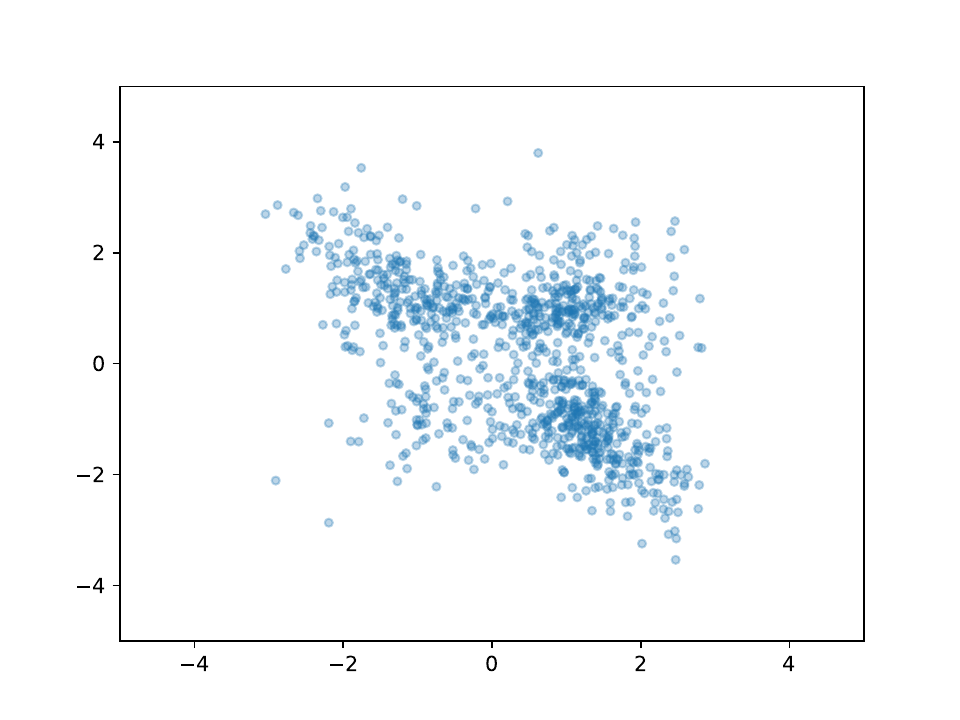} 
        \caption{Monte Carlo}
    \end{subfigure}
    \begin{subfigure}{0.12\textwidth}
        \centering
        \cropfigsole{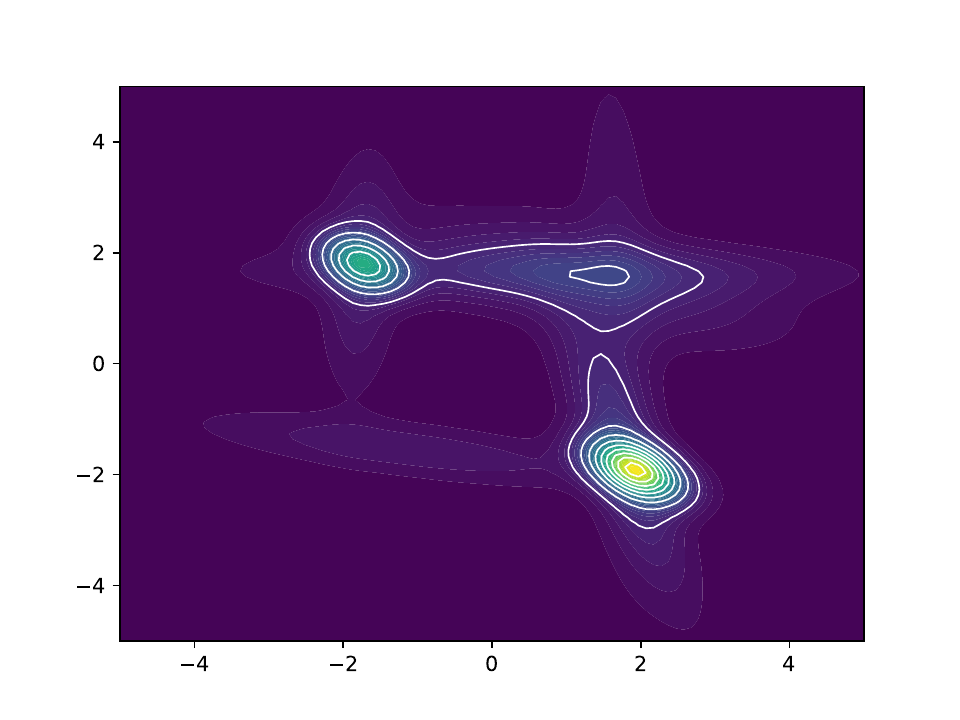}
        \caption{EKF}
    \end{subfigure}
    \begin{subfigure}{0.12\textwidth}
        \centering
        \cropfigsole{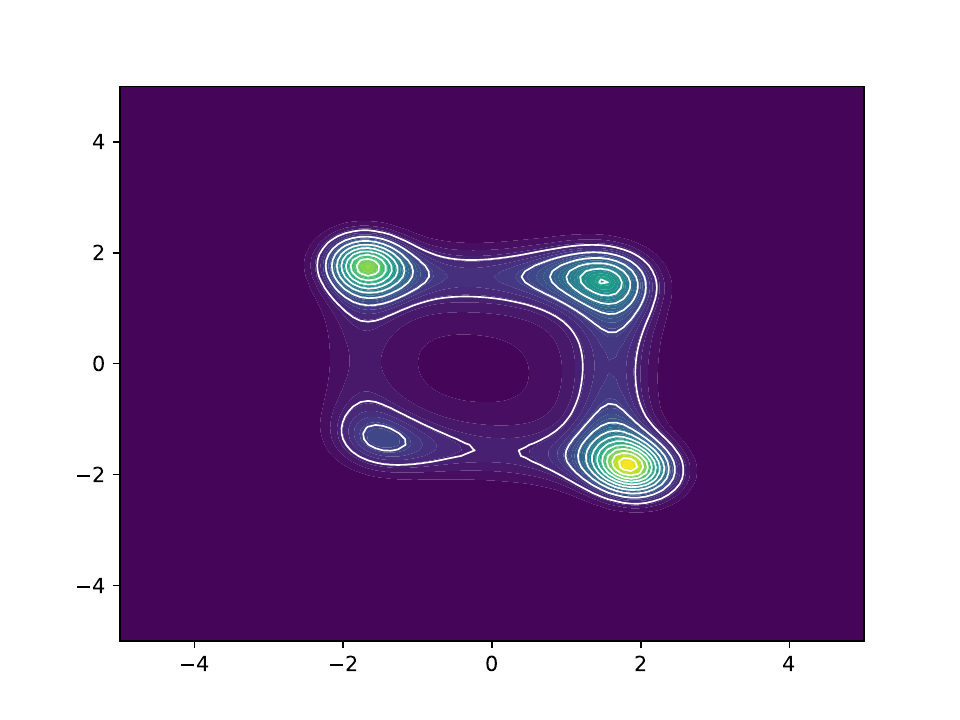}
        \caption{GridGMM}
    \end{subfigure}
    \begin{subfigure}{0.12\textwidth}
        \centering
        \cropfigsole{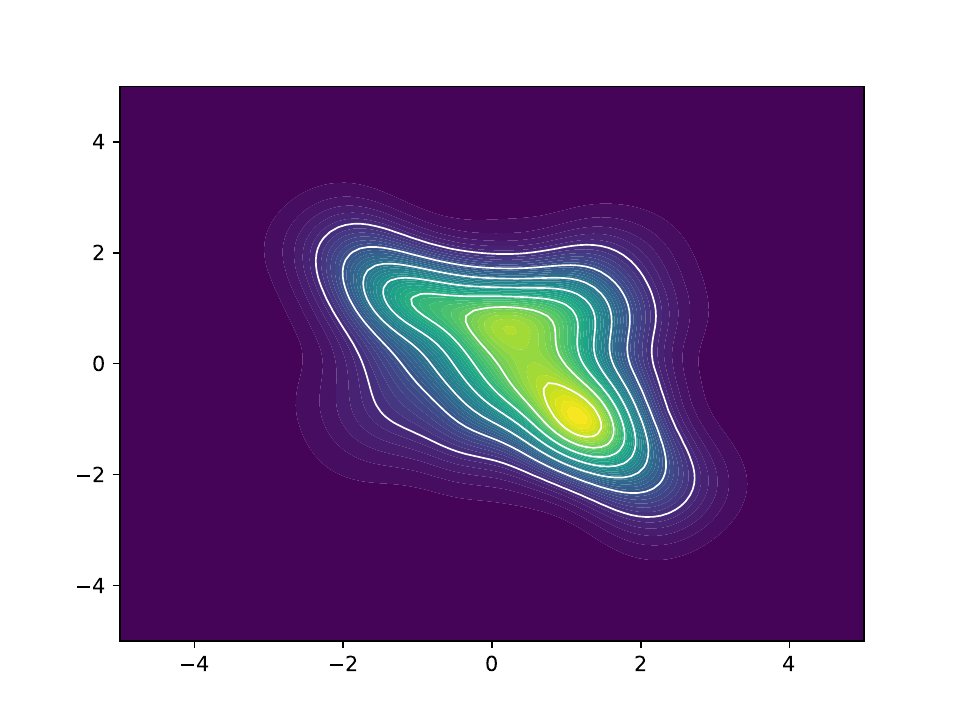}
        \caption{BNF}
    \end{subfigure}
    \caption{Visual Comparison for the computed belief at time $k=9$ of the 
    (a)-(d) Van der Pol
    and
    (e)-(h) stable oscillator systems.}
    \label{fig: belief step9}
\end{figure*}

\paragraph{Experimental Setup} Experiments were performed on two highly non-linear latent stochastic systems: Van der Pol with additive Gaussian noise and a stable oscillator with multiplicative non-Gaussian noise. The latent initial state distribution is a Gaussian distribution. For each GMM method, $p(\px_0)$ is learned with 10 mixture components to avoid overfitting to the initial state data. The resolution of the grid method is 20-by-20. Each comparison uses a Multitask Gaussian Process Regression with a radial basis function kernel. For BNF, we used Bernstein polynomials with degrees 10, 20, and 30 
for each component $g_i$.
Each method is trained with 1K initial state data points 10K state-transition data points. Prediction accuracy is evaluated using average log-likelihood \eqref{eq:opt_obj} on a large test data set (separate from the training data set), consisting of Monte-Carlo samples from the true system.


Fig.~\ref{fig: belief step9} shows a visual comparison of the computed beliefs at time step 
$k=9$ for the Van der Pol system (Fig.~\ref{fig: belief step9}a–d) and the stable oscillator (Fig.~\ref{fig: belief step9}e–h). Figs.~\ref{fig:add_gaus_comp}–\ref{fig:non_add_gaus_comp} present the average log-likelihood results. Overall, we observe that BNF performs comparably to the grid-based method in the additive Gaussian case and significantly outperforms all baselines under non-Gaussian noise. Additionally, performance improves with increasing polynomial degree.

\subsubsection{Additive Gaussian Noise System}
\begin{figure}[t] 
    \centering
    \includegraphics[trim=10 0 0 40, clip, width=0.5\textwidth]{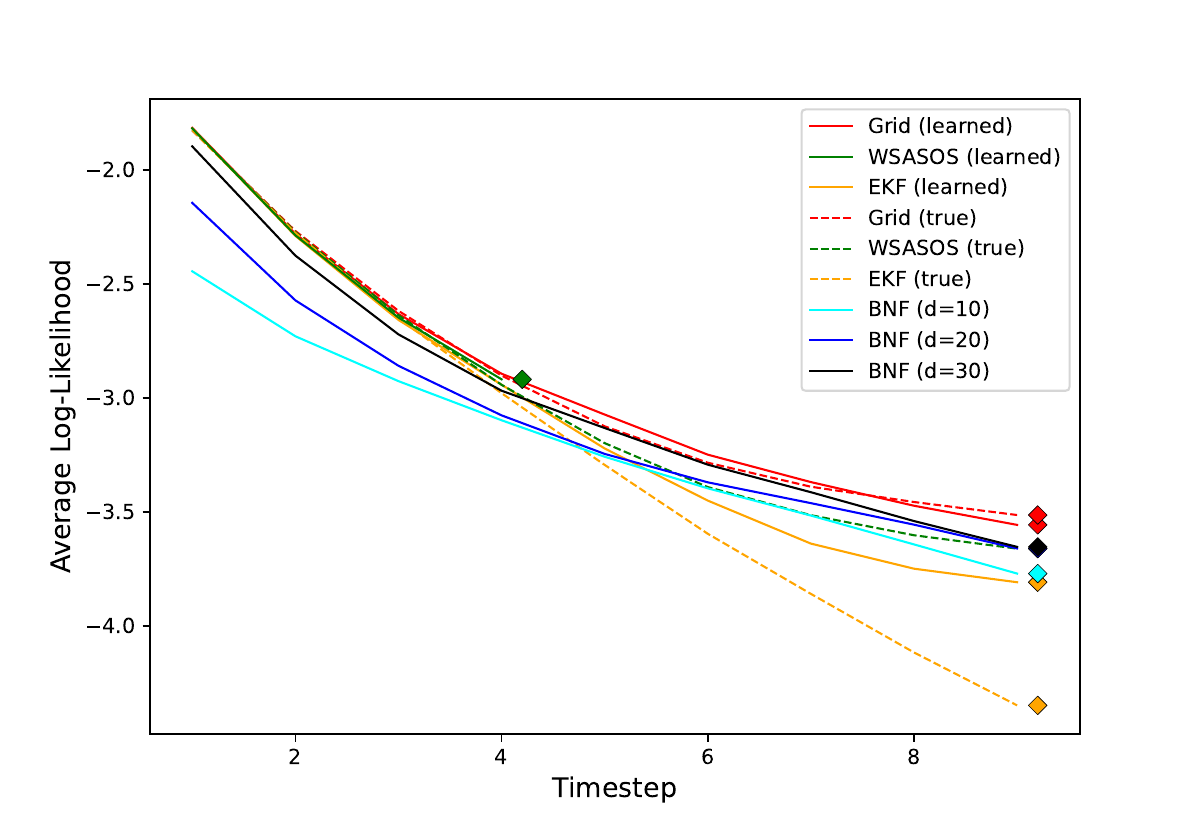}
    \caption{\textbf{Additive-Gaussian Noise} - Belief accuracy 
    }
    \label{fig:add_gaus_comp}
\end{figure}

As can be seen in Fig.~\ref{fig:add_gaus_comp}, BNF generates a more accurate prediction than EKF, however, the grid method performs the best. Since the system is additive Gaussian, GP regression learns a \textit{very} accurate model. Since the grid method produces the least propagation error among the GMM methods, it is able to achieve a very accurate prediction. WSASOS (learned) times out after 6 time steps since the number of components in the GMM explodes exponentially. 

To isolate \textit{propagation error} from \textit{learning error}, we compared BNF to baseline methods using the true underlying system dynamics, effectively isolating only the error accumulated via propagation. Alternatively, since BNF propagation is exact, the induced error is \textit{only} learning error. As can be seen, BNF performs comparably despite starting from a more inaccurate initial distribution.


In BNF, the effect of the degree of the polynomial
is most notably observed the estimate of the initial distribution. The initial distribution has relatively small covariance, thus, the polynomial must be able to attain a steep transition, which requires polynomials of higher degree.

\subsubsection{Multiplicative Non-Gaussian Noise System}
\begin{figure}[t] 
    \centering
    \includegraphics[trim=10 0 0 40, clip, width=0.5\textwidth]{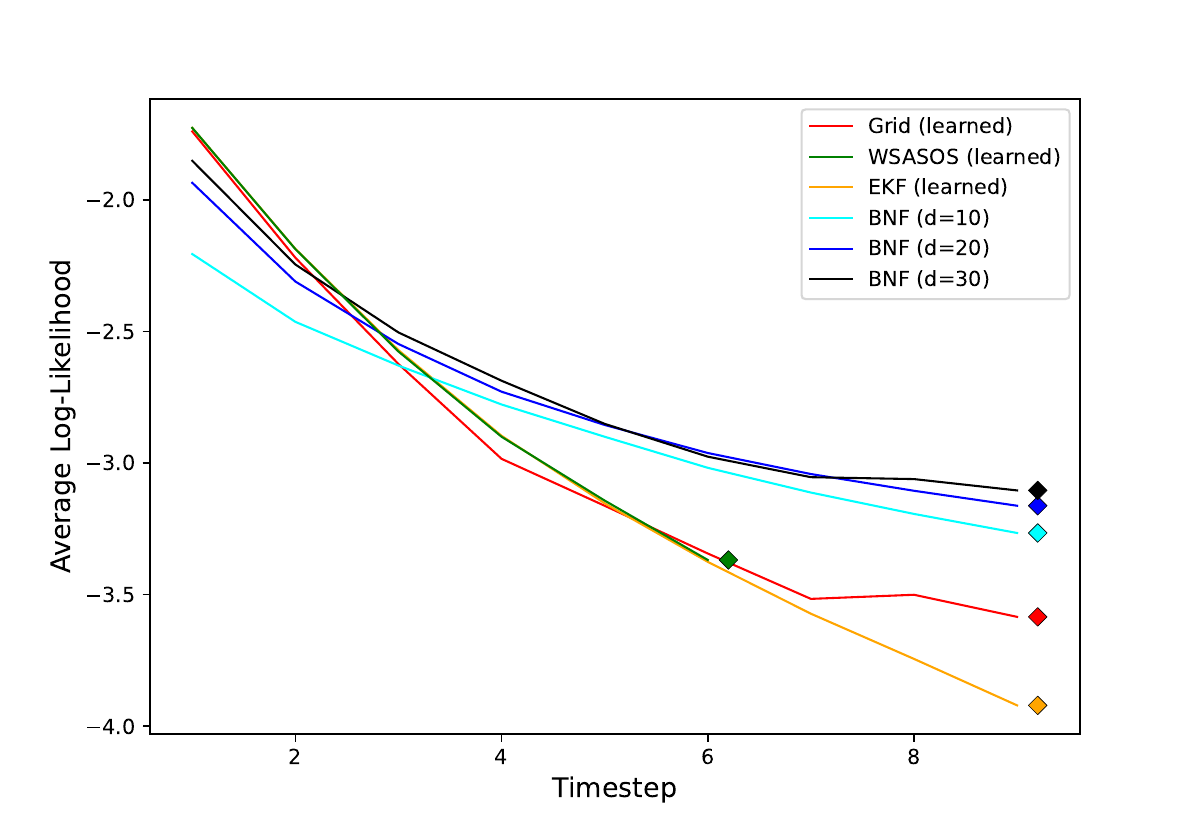}
    \caption{\textbf{Multipli. Non-Gaussian Noise} - Belief accuracy }
    \label{fig:non_add_gaus_comp}
\end{figure}

Fig. \ref{fig:non_add_gaus_comp} illustrates the universal modeling capability of BNFs. Since the true underlying system has highly non-linear non-Gaussian noise, the GP regression models become unable to truly capture the stochastic nature of the system, despite having ample data. Consequently, unlike the additive-Gaussian case, the GMM methods now incur significant propagation \textit{and} learning error. Conversely, BNF is able to more accurately capture the complex stochasticity in the true system (visually seen in Fig. \ref{fig: belief step9} (f-g)). In particular, the inaccuracy in the stochastic model for GP-based collapses the belief over the stable points of the system, whereas BNF maintains a better (more uncertain) representation of the belief. BNF exhibits a consistent trend with the previous experiment when varying the degree. This experiment demonstrates the power of shifting all prediction error to only learning error: prediction accuracy is solely a function of the representational capability of the model and the methods used for training.


These experiments illustrate the power of BNF in learning and propagating PDFs of complex, realistic systems. Unlike baseline methods, BNF maintains high predictive fidelity even under non-Gaussian noise. 
Given BNF's exact propagation property, its sole error is rooted in learning, which can be mitigated using higher polynomial degrees.

\section{Conclusion}
In this work, we introduced a Bernstein Normalizing Flows (BNFs), a novel model class that enables universal PDF approximation and exact belief propagation by combining Bernstein polynomials with normalizing flows. BNFs eliminate approximation error while supporting efficient sampling and evaluation. 
We provided the theoretical foundations for this approach and empirically demonstrate that BNFs are particularly effective in nonlinear, non-Gaussian settings, where state-of-the-art methods suffer. 


\bibliography{bibliography}

\ifarxiv
\else
    \makeatletter
\@ifundefined{isChecklistMainFile}{
  \newif\ifreproStandalone
  \reproStandalonetrue
}{
  \newif\ifreproStandalone
  \reproStandalonefalse
}
\makeatother

\ifreproStandalone
\documentclass[letterpaper]{article}
\usepackage[submission]{aaai2026}
\setlength{\pdfpagewidth}{8.5in}
\setlength{\pdfpageheight}{11in}
\usepackage{times}
\usepackage{helvet}
\usepackage{courier}
\usepackage{xcolor}
\frenchspacing

\begin{document}
\fi
\setlength{\leftmargini}{20pt}
\makeatletter\def\@listi{\leftmargin\leftmargini \topsep .5em \parsep .5em \itemsep .5em}
\def\@listii{\leftmargin\leftmarginii \labelwidth\leftmarginii \advance\labelwidth-\labelsep \topsep .4em \parsep .4em \itemsep .4em}
\def\@listiii{\leftmargin\leftmarginiii \labelwidth\leftmarginiii \advance\labelwidth-\labelsep \topsep .4em \parsep .4em \itemsep .4em}\makeatother

\setcounter{secnumdepth}{0}
\renewcommand\thesubsection{\arabic{subsection}}
\renewcommand\labelenumi{\thesubsection.\arabic{enumi}}

\newcounter{checksubsection}
\newcounter{checkitem}[checksubsection]

\newcommand{\checksubsection}[1]{%
  \refstepcounter{checksubsection}%
  \paragraph{\arabic{checksubsection}. #1}%
  \setcounter{checkitem}{0}%
}

\newcommand{\checkitem}{%
  \refstepcounter{checkitem}%
  \item[\arabic{checksubsection}.\arabic{checkitem}.]%
}
\newcommand{\question}[2]{\normalcolor\checkitem #1 #2 \color{blue}}
\newcommand{\ifyespoints}[1]{\makebox[0pt][l]{\hspace{-15pt}\normalcolor #1}}

\section*{Reproducibility Checklist}

\vspace{1em}
\hrule
\vspace{1em}

\textbf{Instructions for Authors:}

This document outlines key aspects for assessing reproducibility. Please provide your input by editing this \texttt{.tex} file directly.

For each question (that applies), replace the ``Type your response here'' text with your answer.

\vspace{1em}
\noindent
\textbf{Example:} If a question appears as
\begin{center}
\noindent
\begin{minipage}{.9\linewidth}
\ttfamily\raggedright
\string\question \{Proofs of all novel claims are included\} \{(yes/partial/no)\} \\
Type your response here
\end{minipage}
\end{center}
you would change it to:
\begin{center}
\noindent
\begin{minipage}{.9\linewidth}
\ttfamily\raggedright
\string\question \{Proofs of all novel claims are included\} \{(yes/partial/no)\} \\
yes
\end{minipage}
\end{center}
Please make sure to:
\begin{itemize}\setlength{\itemsep}{.1em}
\item Replace ONLY the ``Type your response here'' text and nothing else.
\item Use one of the options listed for that question (e.g., \textbf{yes}, \textbf{no}, \textbf{partial}, or \textbf{NA}).
\item \textbf{Not} modify any other part of the \texttt{\string\question} command or any other lines in this document.\\
\end{itemize}

You can \texttt{\string\input} this .tex file right before \texttt{\string\end\{document\}} of your main file or compile it as a stand-alone document. Check the instructions on your conference's website to see if you will be asked to provide this checklist with your paper or separately.

\vspace{1em}
\hrule
\vspace{1em}

The questions start here

\checksubsection{General Paper Structure}
\begin{itemize}

\question{Includes a conceptual outline and/or pseudocode description of AI methods introduced}{(yes/partial/no/NA)}
yes

\question{Clearly delineates statements that are opinions, hypothesis, and speculation from objective facts and results}{(yes/no)}
yes

\question{Provides well-marked pedagogical references for less-familiar readers to gain background necessary to replicate the paper}{(yes/no)}
yes

\end{itemize}
\checksubsection{Theoretical Contributions}
\begin{itemize}

\question{Does this paper make theoretical contributions?}{(yes/no)}
yes

	\ifyespoints{\vspace{1.2em}If yes, please address the following points:}
        \begin{itemize}
	
	\question{All assumptions and restrictions are stated clearly and formally}{(yes/partial/no)}
	partial

	\question{All novel claims are stated formally (e.g., in theorem statements)}{(yes/partial/no)}
	yes

	\question{Proofs of all novel claims are included}{(yes/partial/no)}
	yes

	\question{Proof sketches or intuitions are given for complex and/or novel results}{(yes/partial/no)}
	yes

	\question{Appropriate citations to theoretical tools used are given}{(yes/partial/no)}
	yes

	\question{All theoretical claims are demonstrated empirically to hold}{(yes/partial/no/NA)}
	partial

	\question{All experimental code used to eliminate or disprove claims is included}{(yes/no/NA)}
	yes
	
	\end{itemize}
\end{itemize}

\checksubsection{Dataset Usage}
\begin{itemize}

\question{Does this paper rely on one or more datasets?}{(yes/no)}
no

\ifyespoints{If yes, please address the following points:}
\begin{itemize}

	\question{A motivation is given for why the experiments are conducted on the selected datasets}{(yes/partial/no/NA)}
	NA

	\question{All novel datasets introduced in this paper are included in a data appendix}{(yes/partial/no/NA)}
	NA

	\question{All novel datasets introduced in this paper will be made publicly available upon publication of the paper with a license that allows free usage for research purposes}{(yes/partial/no/NA)}
	NA

	\question{All datasets drawn from the existing literature (potentially including authors' own previously published work) are accompanied by appropriate citations}{(yes/no/NA)}
	NA

	\question{All datasets drawn from the existing literature (potentially including authors' own previously published work) are publicly available}{(yes/partial/no/NA)}
	NA

	\question{All datasets that are not publicly available are described in detail, with explanation why publicly available alternatives are not scientifically satisficing}{(yes/partial/no/NA)}
	NA

\end{itemize}
\end{itemize}

\checksubsection{Computational Experiments}
\begin{itemize}

\question{Does this paper include computational experiments?}{(yes/no)}
yes

\ifyespoints{If yes, please address the following points:}
\begin{itemize}

	\question{This paper states the number and range of values tried per (hyper-) parameter during development of the paper, along with the criterion used for selecting the final parameter setting}{(yes/partial/no/NA)}
	yes

	\question{Any code required for pre-processing data is included in the appendix}{(yes/partial/no)}
	yes

	\question{All source code required for conducting and analyzing the experiments is included in a code appendix}{(yes/partial/no)}
	yes

	\question{All source code required for conducting and analyzing the experiments will be made publicly available upon publication of the paper with a license that allows free usage for research purposes}{(yes/partial/no)}
	yes
        
	\question{All source code implementing new methods have comments detailing the implementation, with references to the paper where each step comes from}{(yes/partial/no)}
	partial

	\question{If an algorithm depends on randomness, then the method used for setting seeds is described in a way sufficient to allow replication of results}{(yes/partial/no/NA)}
	yes

	\question{This paper specifies the computing infrastructure used for running experiments (hardware and software), including GPU/CPU models; amount of memory; operating system; names and versions of relevant software libraries and frameworks}{(yes/partial/no)}
	no

	\question{This paper formally describes evaluation metrics used and explains the motivation for choosing these metrics}{(yes/partial/no)}
	yes

	\question{This paper states the number of algorithm runs used to compute each reported result}{(yes/no)}
	yes

	\question{Analysis of experiments goes beyond single-dimensional summaries of performance (e.g., average; median) to include measures of variation, confidence, or other distributional information}{(yes/no)}
	no

	\question{The significance of any improvement or decrease in performance is judged using appropriate statistical tests (e.g., Wilcoxon signed-rank)}{(yes/partial/no)}
	no

	\question{This paper lists all final (hyper-)parameters used for each model/algorithm in the paper’s experiments}{(yes/partial/no/NA)}
	yes

\end{itemize}
\end{itemize}
\ifreproStandalone
\end{document}
\fi
\fi

\appendix
\section{Appendix}

\subsection{Choosing $\Omega$}
Choosing $\Omega$ for a given problem can affect the ability for the BNF model learn an accurate representation of the data. Intuitively, since BNF models are polynomials, the approximation of capability is primarily limited by the degree of smoothness of the underlying density. For instance, an underlying PDF that has a sharp peak requires a high degree BNF to model it properly. When transforming $\reals^n$ to $\uspace^n$, we can choose $\Omega$ to try to reduce the sharpness of the underlying data distribution. As an illustrative example, suppose BNF is used to model a roughly independent Gaussian density $p^\star(\px)$ with small covariance. If we choose $\Omega$ to be the CDF of a Gaussian distribution that closely approximates $p^\star(\px)$, then $p(\pu)$ will be roughly uniform.

\paragraph{Moment Matching Transform} For the experiments in this work, we use Gaussian CDFs for the components of $\Omega$, defined by two parameters, a mean and a variance. To determine the mean and variance of the CDF, we calculated the mean and variance for each state-component (independently) across all training data. Additionally, for scenarios with steep peaks far away from the mean, we manually enlarged the variance beyond the variance of the data used for the components of $\Omega$. Figure \ref{fig:omega_comparison} illustrates the difference between a poorly chosen $\Omega$ and one selected using the moment matching technique for a simple density estimation task. The poorly tuned $\Omega$ forces the $\pu$-density to have very steep peaks along the boundary of $\uspace^n$, resulting in a less accurate density estimate than the moment-matched $\Omega$.

\newcommand{\cropfigsoleomega}[1]{
    \includegraphics[width=\textwidth, trim=95 30 55 40, clip]{#1}%
}


\begin{figure*}[ht]
    \centering

    \begin{minipage}[c]{0.38\textwidth}
        \centering
        \vspace{15mm}
        \cropfigsoleomega{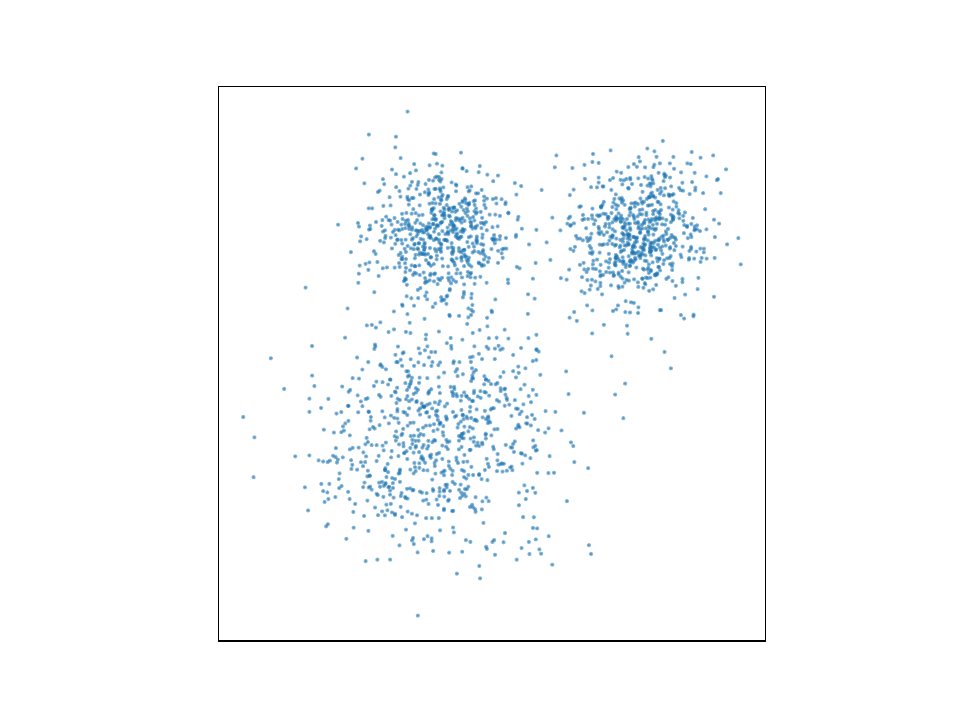}
        \caption*{$\reals^n$ data}
        \label{fig:true_data}
    \end{minipage}
    \begin{minipage}[t]{0.6\textwidth}
        \centering
        \begin{minipage}[t]{\textwidth}
            \centering
            \begin{subfigure}[t]{0.32\textwidth}
                \centering
                \cropfigsoleomega{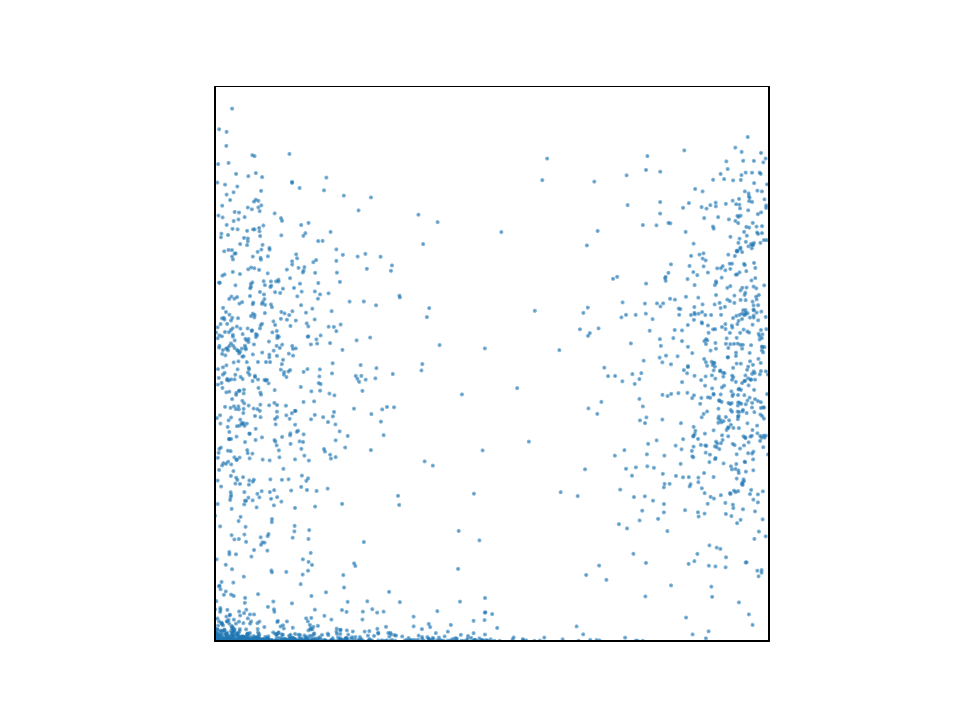}
                \caption{$\uspace^n$ data}
            \end{subfigure}
            \begin{subfigure}[t]{0.32\textwidth}
                \centering
                \cropfigsoleomega{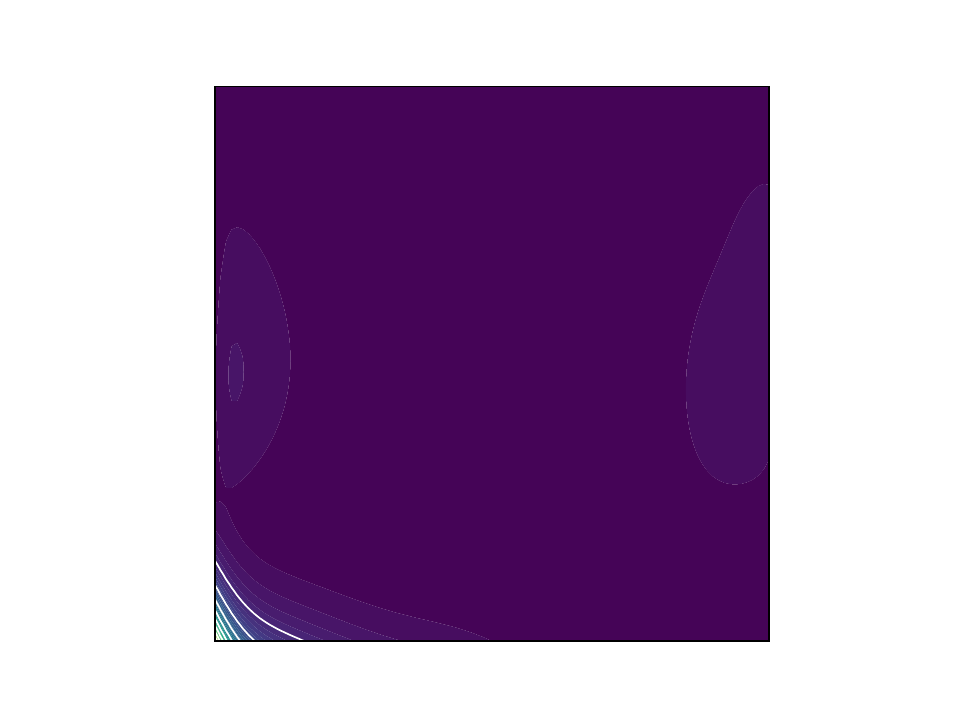}
                \caption{Learned $p(\pu)$}
            \end{subfigure}
            \begin{subfigure}[t]{0.32\textwidth}
                \centering
                \cropfigsoleomega{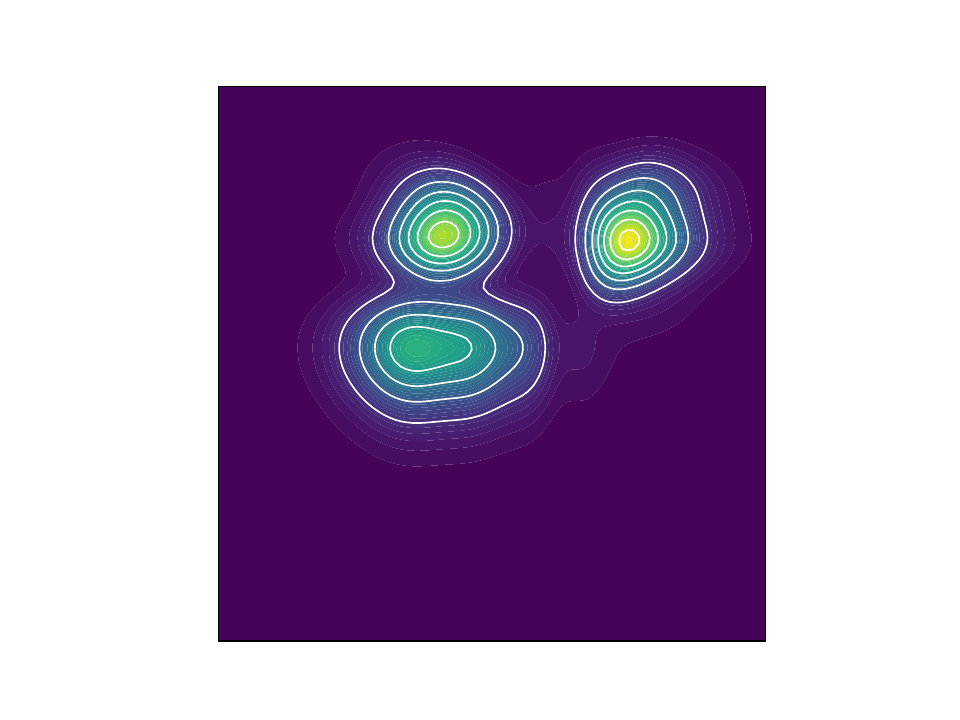}
                \caption{Learned $p(\px)$}
            \end{subfigure}
        \end{minipage}

        \begin{minipage}[t]{\textwidth}
            \centering
            \begin{subfigure}[t]{0.32\textwidth}
                \centering
                \cropfigsoleomega{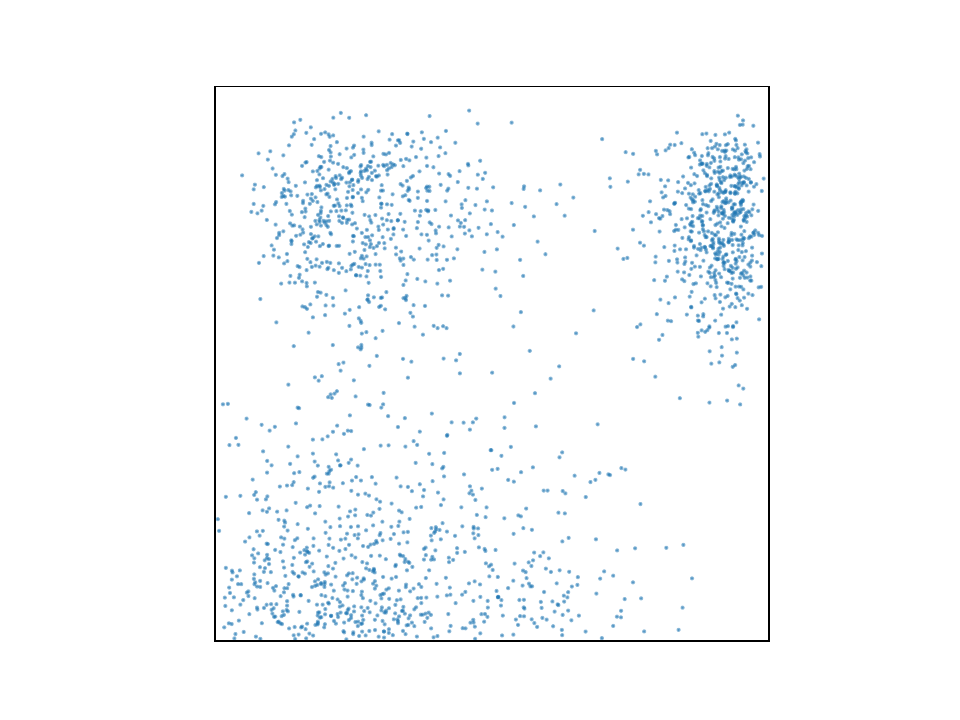}
                \caption{$\uspace^n$ data}
            \end{subfigure}
            \begin{subfigure}[t]{0.32\textwidth}
                \centering
                \cropfigsoleomega{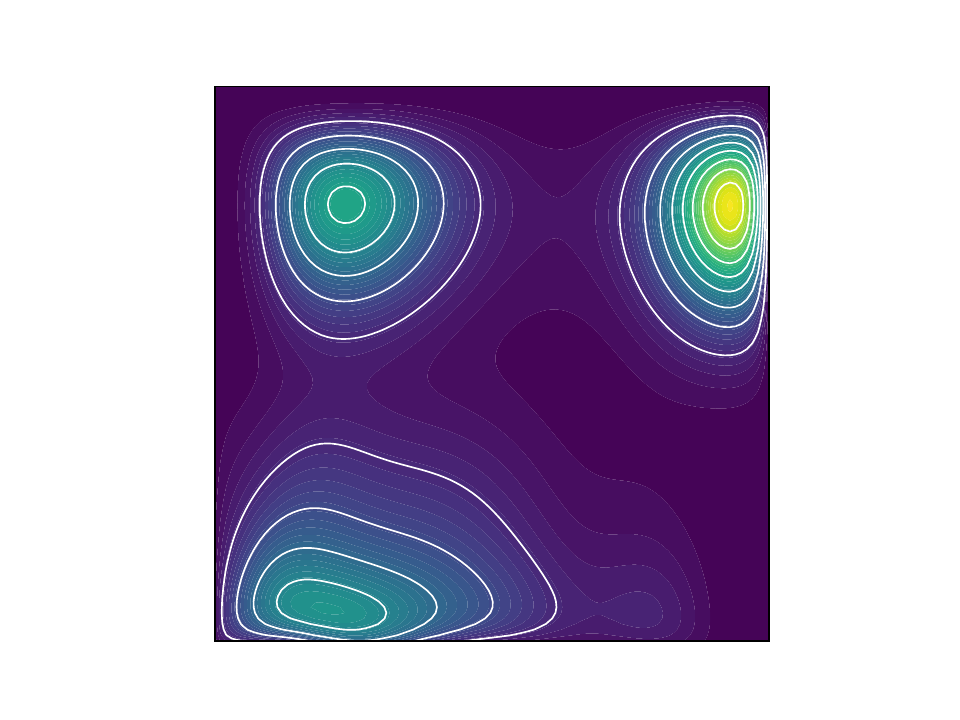}
                \caption{Learned $p(\pu)$}
            \end{subfigure}
            \begin{subfigure}[t]{0.32\textwidth}
                \centering
                \cropfigsoleomega{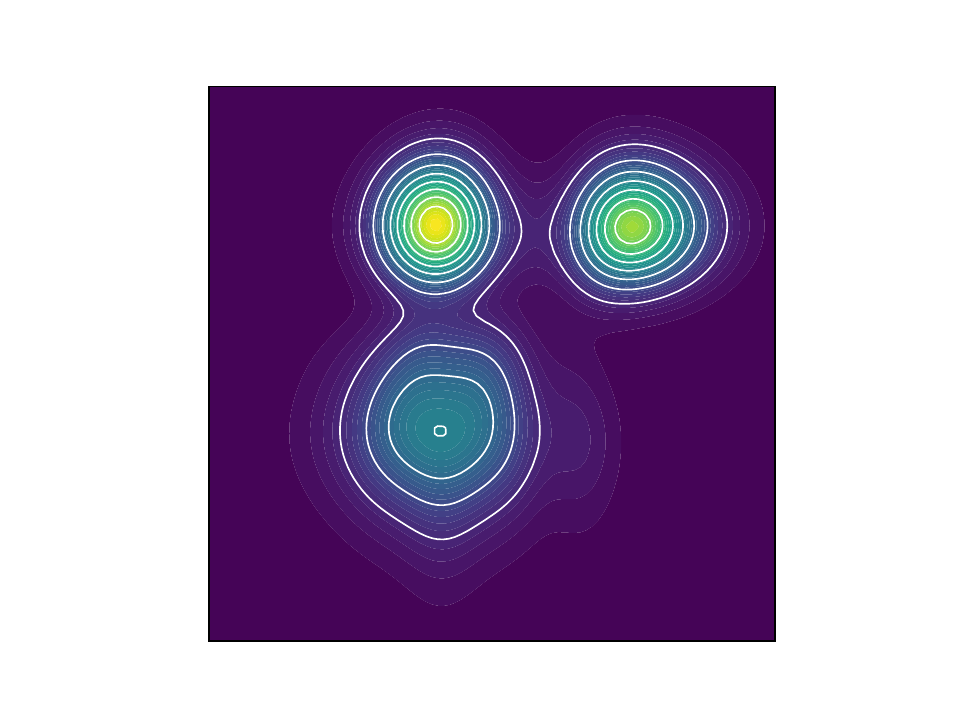}
                \caption{Learned $p(\px)$}
            \end{subfigure}
        \end{minipage}
    \end{minipage}
    \hfill

    \caption{Comparison between poorly selected $\Omega$ (top row) and moment-matched $\Omega$ (bottom row) on a simple density estimation task using a degree $15$ BNF. The poorly chosen $\Omega$ causes the $\uspace^n$ data (a) to be concentrated on the edges, creating steep density transitions (b), and thus an inaccurate estimate (c-d). The moment-matched $\Omega$ has a more even distribution in $\uspace^n$ (d), which reduces the sharpness of the pdf (e), leading to a more accurate estimate (e-f).}
    \label{fig:omega_comparison}
\end{figure*}


\subsection{Proof of Lemma \ref{lem:normalize}}
\begin{proof}
    The integral of any 1D Bernstein basis function of degree $d$ over $[0,1]$ is known to be
    \begin{equation} \label{eq:basis_int}
        \int_0^1 \phi_j^d(u) du = \frac{1}{n+1}.
    \end{equation}
    Consider the multivariate 
    Bernstein polynomial
    \begin{equation}
        \pi(\pu) = \sum_{\bj=\mathbf{0}}^{\bd} b_{\bj} \phi_\bj^\bd(\pu),
    \end{equation}
    with a $n$-dimensional coefficient tensor $\pb$. Without loss of generality, we can integrate over $u_n$
    \begin{align}
        \int_0^1 \pi(\pu) du_n = \int_0^1 \sum_{j_1 = 0}^{d_1} \cdots \sum_{j_{n}=0}^{d_{n}} b_{\bj} \prod_{l=1}^n \phi_{j_l}^{d_l}(u_l) du_n \notag \\
         = \sum_{j_1 = 0}^{d_1} \cdots \sum_{j_{n-1}=0}^{d_{n-1}} \Big(\prod_{l=1}^{n-1} \phi_{j_l}^{d_l}(u_l) \Big) \sum_{j_{n}=0}^{d_{n}} b_{\bj} \int_0^1 \phi_{j_n}^{d_n}(u_n) du_n \notag \\
         = \sum_{j_1 = 0}^{d_1} \cdots \sum_{j_{n-1}=0}^{d_{n-1}} \Big(\prod_{l=1}^{n-1} \phi_{j_l}^{d_l}(u_l) \Big) \sum_{j_{n}=0}^{d_{n}} b_{\bj} \frac{1}{1+d_n}.
    \end{align}
    Therefore, $\int_0^1 \pi(\pu) du_n = 1$ iff 
    \begin{align}
        \sum_{j_{n}=0}^{d_{n}} b_{j_n} \frac{1}{1+d_n} &= 1.
    \end{align}
\end{proof}
\subsection{Proof of Theorem \ref{thm:universal}}
\begin{proof}
    We begin by recalling the definitional properties of a probability divergence \cite{amari2016information}. We consider divergence between arbitrary distributions $p\in \mathcal{P}$ parameterized by (generally) infinite-dimensional parameter vectors $\xi$ and distributions $\mathcal{P}_\theta$ parameterized with finite parameter vectors $\theta$. A probability divergence $\mathcal{D}(\cdot \| \cdot)$ is defined as a function $\mathcal{D} : \mathcal{P} \times \mathcal{P}_\theta \rightarrow \reals$ that must satisfy the following three properties for any $p_\xi \in \mathcal{P}, p_\theta \in \mathcal{P}_\theta$:
    \begin{enumerate}
        \item $\mathcal{D}(p_\xi \| p_\theta) \geq 0$, \label{cond:div_1}
        \item $\mathcal{D}(p_\xi \| p_\theta) = 0$ iff $p_\xi = p_\theta$, and \label{cond:div_2}
        \item when $p_\xi$ and $p_\theta$ are sufficiently close, denoted $\xi = \theta + d\theta$, the taylor expansion of $\mathcal{D}$ is
        \begin{equation} \label{eq:quad_form}
            \mathcal{D}(p_\theta \| p_{\theta + d\theta}) = \frac{1}{2} \sum \lambda_{i, j}(\theta) d\theta_i d\theta_j + O(|d\theta|^3)
        \end{equation}
        where $\lambda_{i, j}(\theta)$ forms a positive definite matrix $\Lambda(\theta)$.\label{cond:div_3}
    \end{enumerate}
    Since divergence is minimized only when $p_\xi = p_\theta$ and is differentiable (condition \ref{cond:div_3}), it follows that if, for any $\epsilon > 0$, there exists $\theta$ such that $\mathcal{D}(p_\theta \| p_\xi)$, then for any $\delta > 0$, there exists $\theta$ such that $|p_\theta - p_\xi| < \delta$.
    
    Additionally, the Weierstrass approximation theorem \cite{weierstrass1885analytische} is states that for any real-valued function $f$ on $\uspace^n$ and $\delta > $, there exists a polynomial $\pi(\pu)$ such that for all $\pu \in \uspace^n$, we have $|f(\pu) - \pi(\pu)| < \delta|$.
    
    Consider the factorization of an arbitrary conditional distribution 
    \begin{equation}
        p(\pu | \pw) = p_1(u_1 | \pw) p_2(u_2 | u_1, \pw) \cdots p_n(u_n | \pu_{<n}, \pw).
    \end{equation}
    To prove Theorem \ref{thm:universal}, it suffices to show that there exists a Bernstein polynomial $\pu(\pu)$ with coefficients $\tilde{\pb}$ that i) satisfies the constraints \eqref{eq:opt_const_1} and \eqref{eq:opt_const_2} and ii) can universally approximate an arbitrary factor $p_i$ of $p$, i.e. for any $\epsilon > 0$, $\exists \tilde{\pb}$ such that $\mathcal{D}(\pi \| p_i) < \epsilon$.

    By the Weierstrass theorem, there exists a Bernstein polynomial $\pi^\star(\pu_{\leq i}, \pw)$ with coefficients $\pb^\star$ such that $|p_i(u_i | \pu_{<i}, \pw) - \pi^\star(\pu_{\leq i}, \pw)| < \delta^\star$. With the Bernstein-polynomial proof of the Weierstrass theorem, coefficients of the approximator polynomial are generated by grid evaluations of $p_i(u_i | \pu_{<i}, \pw)$, thereby automatically satisfying \eqref{eq:opt_const_1}, since $p_i >0$ on $\uspace^n$. Then, let $\pi(\pu_{\leq i}, \pw)$ be the normalized version of $\pi^\star$ (satisfying \eqref{eq:opt_const_2}) with Bernstein coefficients $\pb$ such that $\pb_\bj = A \pb^\star$ for some positive normalization tensor $A$. Let $\delta = (\max A) \delta^\star$.
\end{proof}

\subsection{Proof of Theorem \ref{thm:future_class}}


\begin{proof}
    Let
    \begin{align}
        p(\pu' \mid \pu) &= \pi_{\pu'}(\pu', \pu) = \sum_{\bj'}^{\bd'} \sum_{\bj}^{\bd} b_{(\bj', \bj)} \phi_{\bj'}^{\bd'}(\pu')\phi_{\bj}^{\bd}(\pu) \\
        p(\pu) &= \pi_{\pu}(\pu) = \sum_{\bl}^{\bd} a_{\bl} \phi_{\bl}^{\bd}(\pu)
    \end{align}
     
    \begin{align}
        p(\pu') &= \int_{\pu} \pi_{\pu'}(\pu', \pu) \pi_\pu(\pu) d\pu \notag\\
        &= \int_{\pu}\Big[\sum_{\bj'}^{\bd'} \sum_{\bj}^{\bd} b_{(\bj', \bj)} \phi_{\bj'}^{\bd'}(\pu')\phi_{\bj}^{\bd}(\pu) \Big]\Big[\sum_{\bl}^{\bd} a_{\bl} \phi_{\bl}^{\bd}(\pu)\Big]d\pu \notag\\
        &= \sum_{\bj'}^{\bd'} \sum_{\bj}^{\bd} \sum_{\bl}^{\bd} a_{\bl} b_{(\bj', \bj)} \phi_{\bj'}^{\bd'}(\pu') \int_{\pu}  \phi_{\bj}^{\bd}(\pu) \phi_{\bl}^{\bd}(\pu)d\pu \label{eq:marg_deg}
    \end{align}
    where $\int_{\pu}  \phi_{\bj}^{\bd}(\pu) \phi_{\bl}^{\bd}(\pu)d\pu$ is just the integral of the product of Bernstein basis functions over $\uspace^n$ which is a scalar. Therefore, only Bernstein basis functions of degree $\bd'$ remain in \eqref{eq:marg_deg}. Since $\bd'$ is independent of $\bd$, applying \eqref{eq:marg_deg} recursively from $p(\pu_0)$ yields a degree $\bd'$ polynomial for all $p(\pu_k)$ for $k\geq1$.
\end{proof}

\subsection{Experimental Setup Details}
\paragraph{System Dynamics} The Van der Pol system is given by the dynamics
\begin{align}
    x_1' &= x_1 + \Delta t x_2 + v_1\notag\\
    x_2' &= x_2 + \Delta t (\mu (1 - x_1^2) x_2 - x_1) + v_2
\end{align}
where $[v_1, v_2]^T \sim \mathcal{N}([0, 0], 0.1 I)$ and $\Delta t=0.3$. The stable oscillator system is given by the dynamics
\begin{align}
    x_1' &= x_1 + \Delta t (x_1 -x_1^3 - 0.5 x_2) + x_1 v_1 \notag\\
    x_2' &= x_2 + \Delta t (x_2 -x_2^3 - 0.5 x_1) + x_2 v_2
\end{align}
where $[v_1, v_2]^T \sim 0.6 \mathcal{N}([0, 0], \Sigma_1) + 0.4 \mathcal{N}([0.5, 0.5], \Sigma_2)$ and
\[
\Sigma_1 = 
\begin{pmatrix}
0.03 & 0.006 \\
0.006 & 0.03
\end{pmatrix},
\Sigma_2 = 
\begin{pmatrix}
0.03 & -0.006 \\
-0.006 & 0.03
\end{pmatrix}.
\]

\paragraph{Dataset Generation} For each experiment, initial state data was sampled from the initial distribution
\begin{equation}
    \px_0 \sim \mathcal{N}([0.2, 0.1], 0.2 I)
\end{equation}
and state transition data was collected from 1000 sampled 10-timestep trajectories.

\paragraph{Model/Training Parameters}
Each $\Omega$ was constructed with Gaussian CDFs with the mean and variance matched to the mean and variance of all states across both data sets. Additionally, to ease the boundary steepness, a buffer of $2.2$ was added to the variance of the data. This value was determined using rough cross-validation. The initial state models were trained over 3000 epochs with a batch size of 128 and learning rate of 0.01. The state transition models were trained over 150 epochs with a batch size of 1048 with a learning rate 0.1. The initial state model was given a degree increase of 20, and the state-transition model's degree was not increased.

\subsection{BNF Degree Comparison}
\begin{figure}[ht]
    \centering
    \begin{subfigure}[t]{0.48\textwidth}
        \centering
        \includegraphics[trim=10 0 0 40, clip, width=\textwidth]{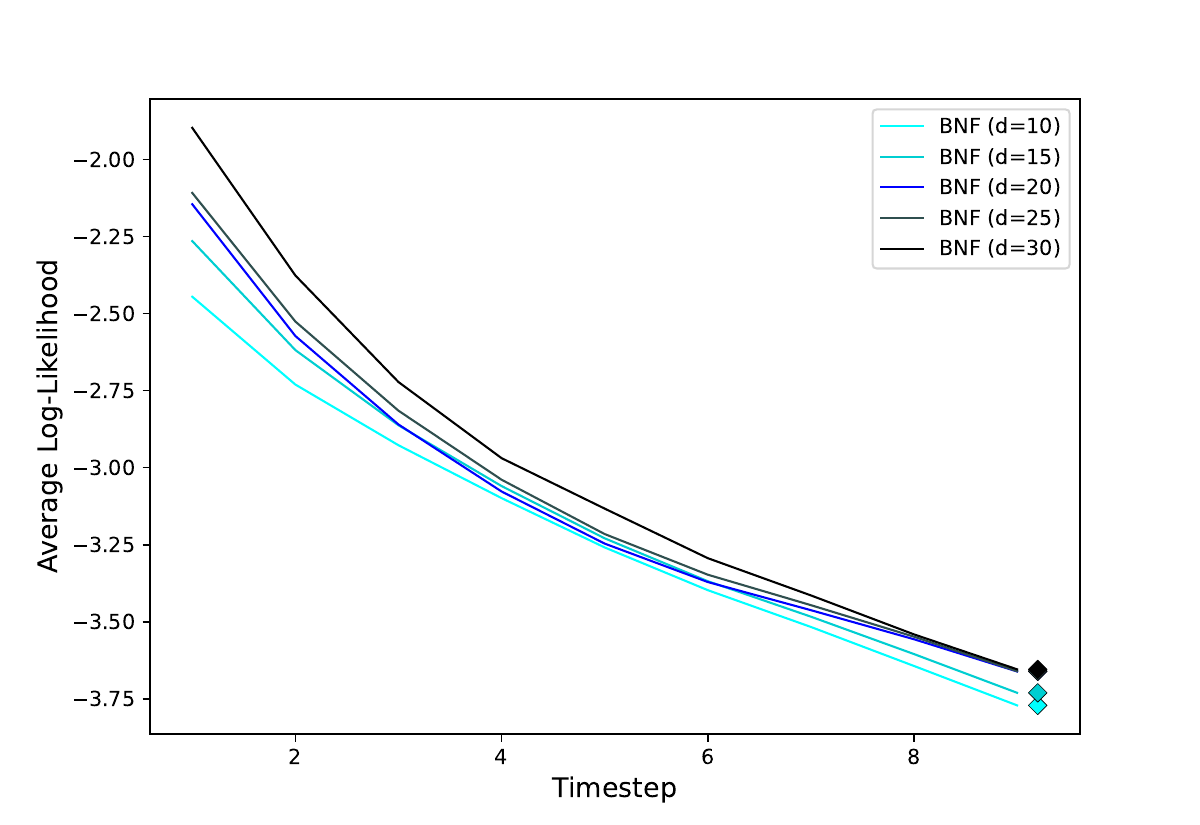}
        \caption{\textbf{Additive-Gaussian Noise} - BNF Degree Comparison}
        \label{fig:deg_comp_ag}
    \end{subfigure}
    \hfill
    \begin{subfigure}[t]{0.48\textwidth}
        \centering
        \includegraphics[trim=10 0 0 40, clip, width=\textwidth]{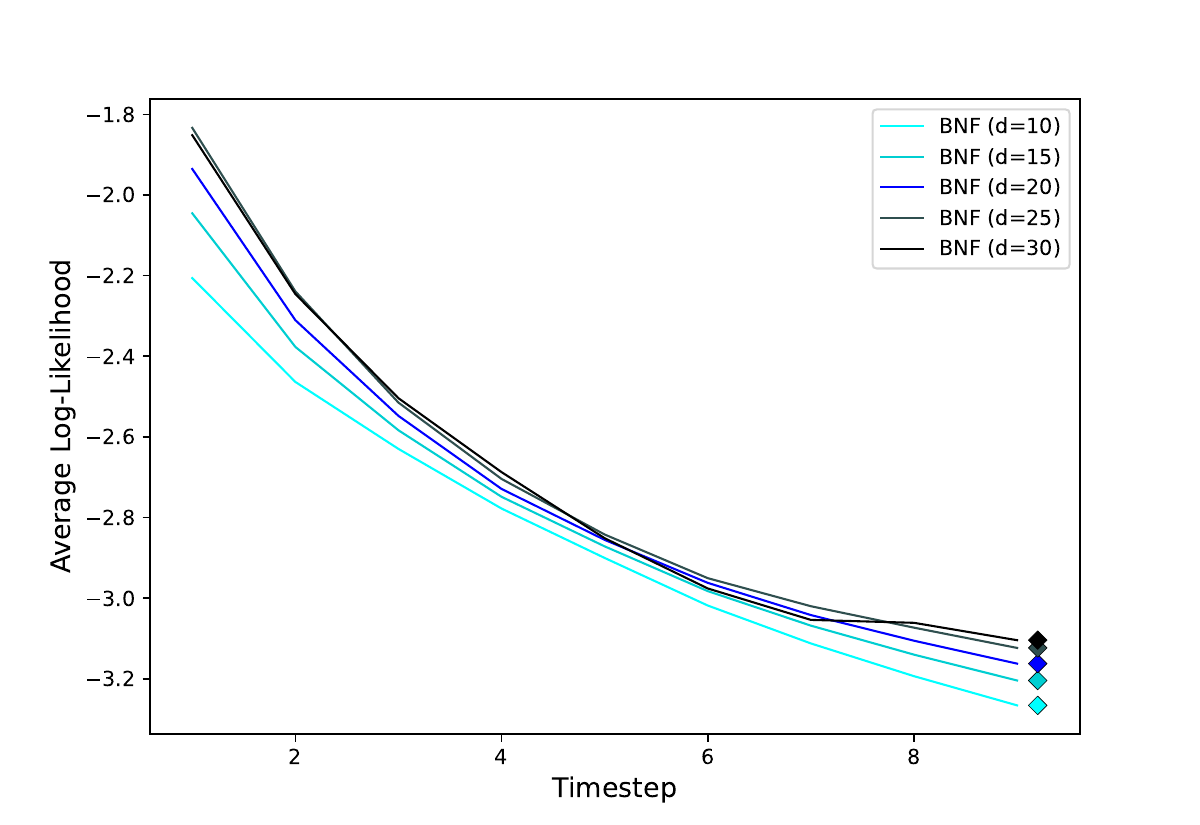}
        \caption{\textbf{Multipli. Non-Gaussian Noise} - BNF Degree Comparison}
        \label{fig:deg_comp_nag}
    \end{subfigure}
    \caption{Degree Comparison}
    \label{fig:deg_comp}
\end{figure}

To further illustrate the convergence of BNF with respect to the choice of degree $\bd$, we provide a more elaborate comparison between BNF results for degrees $\bd=10, 15, 20, 25, 30$. Fig \ref{fig:deg_comp_ag} displays the comparison for the Van der Pol system and Fig. \ref{fig:deg_comp_nag} displays the comparison for the stable oscillator system. As expected, increasing the degree improves prediction accuracy since the learned Markov chain is more accurate.

\section{Incremental Belief Propagation Visual Results}
We provide extended visual results of those shown in Fig. \ref{fig: belief step9} for all time steps $k=1$ to $k=9$ in Figs.~\ref{fig: visual add gaussian 1-9} and \ref{fig: visual non gaussian 1-9}. The grid method used a resolution of $30 \times 30$, and BNF used degree 30. All GMM methods used the learned GP model.

\newcommand{\imgwidth}{0.2\textwidth}
\newcommand{\imgheight}{0.01\textheight}
\newcommand{\cropfig}[1]{
    \includegraphics[width=\imgwidth, trim=55 37 45 20, clip]{#1}%
}

\begin{figure*}[t]
    \centering
    \setlength{\tabcolsep}{0pt}
    \begin{tabular}{c*{5}{c}}
        \rotatebox{90}{\strut \scriptsize Monte Carlo} &
        \cropfig{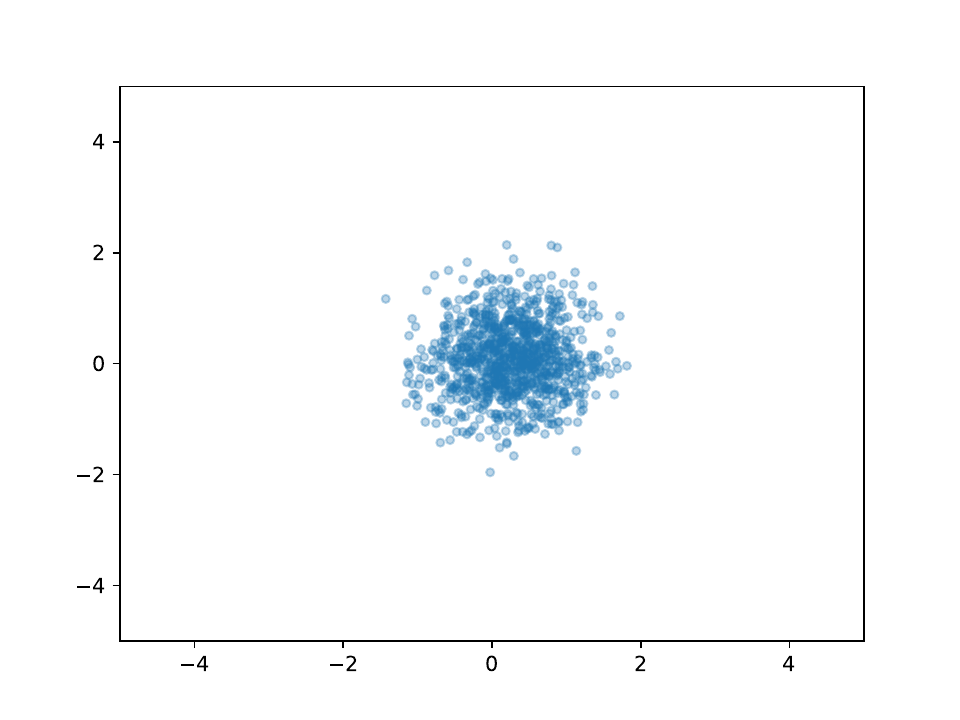} &
        \cropfig{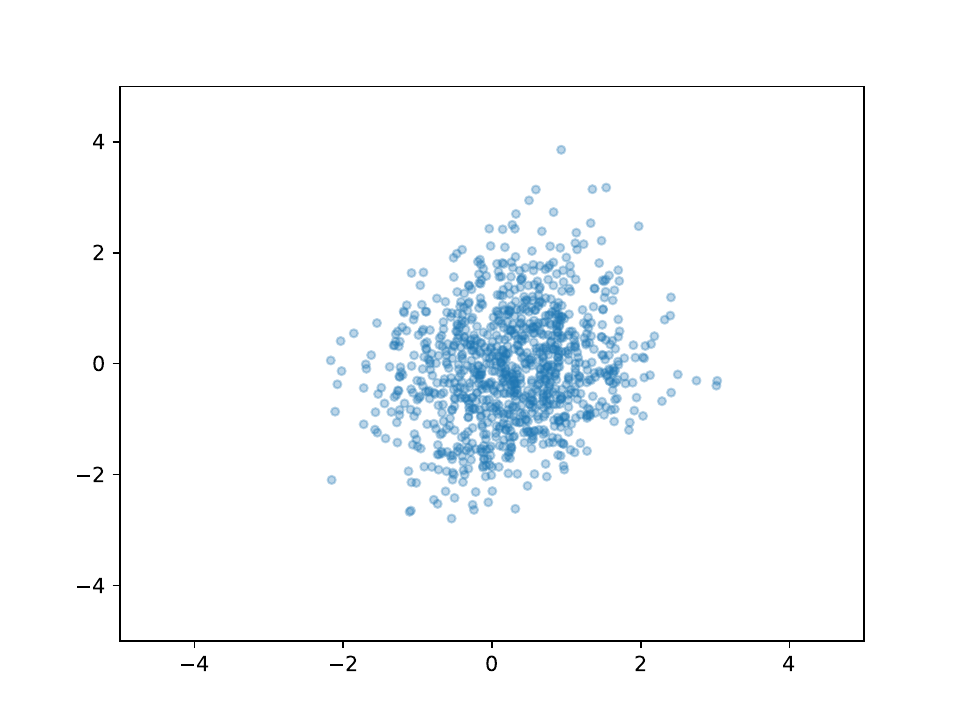} &
        \cropfig{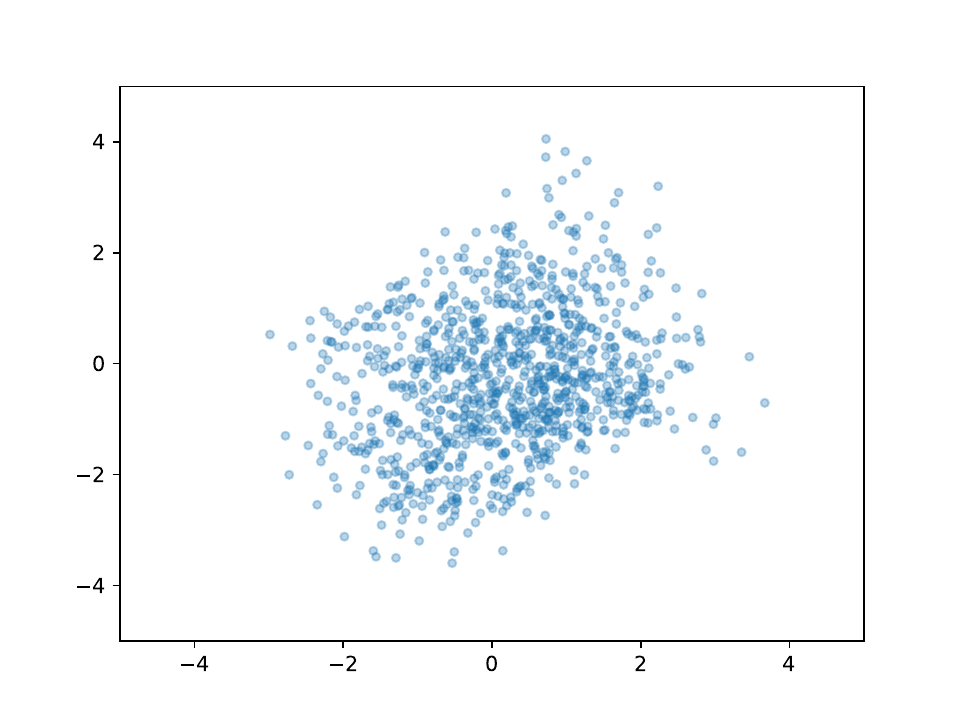} &
        \cropfig{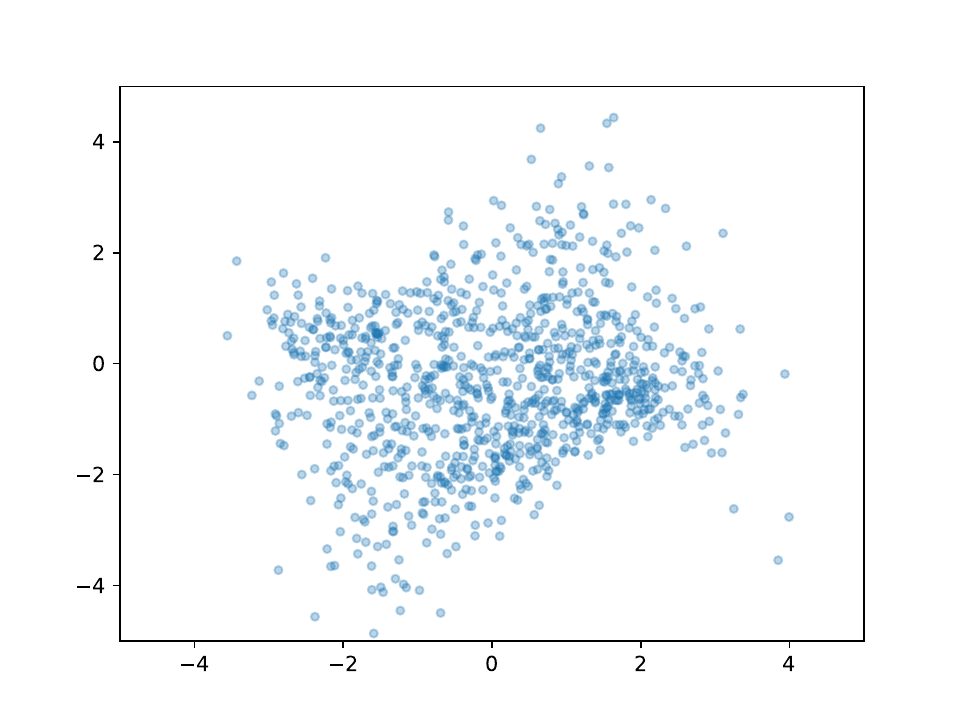} &
        \cropfig{figures/exp_2D_AG/monte_carlo/k_9.pdf} \\

        \rotatebox{90}{\strut \scriptsize EKF} &
        \cropfig{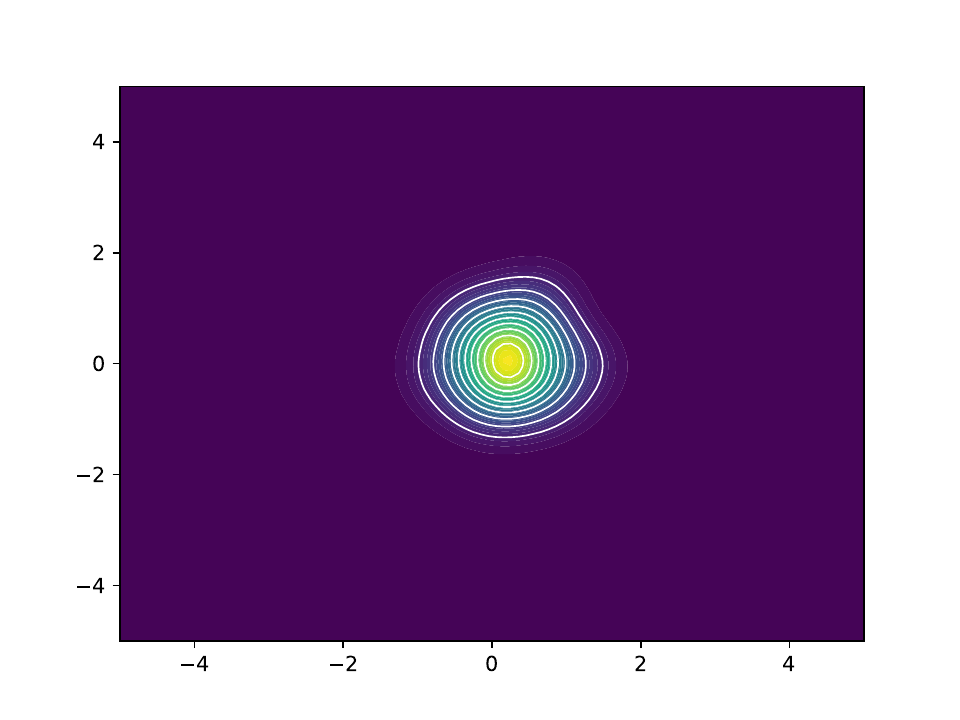} &
        \cropfig{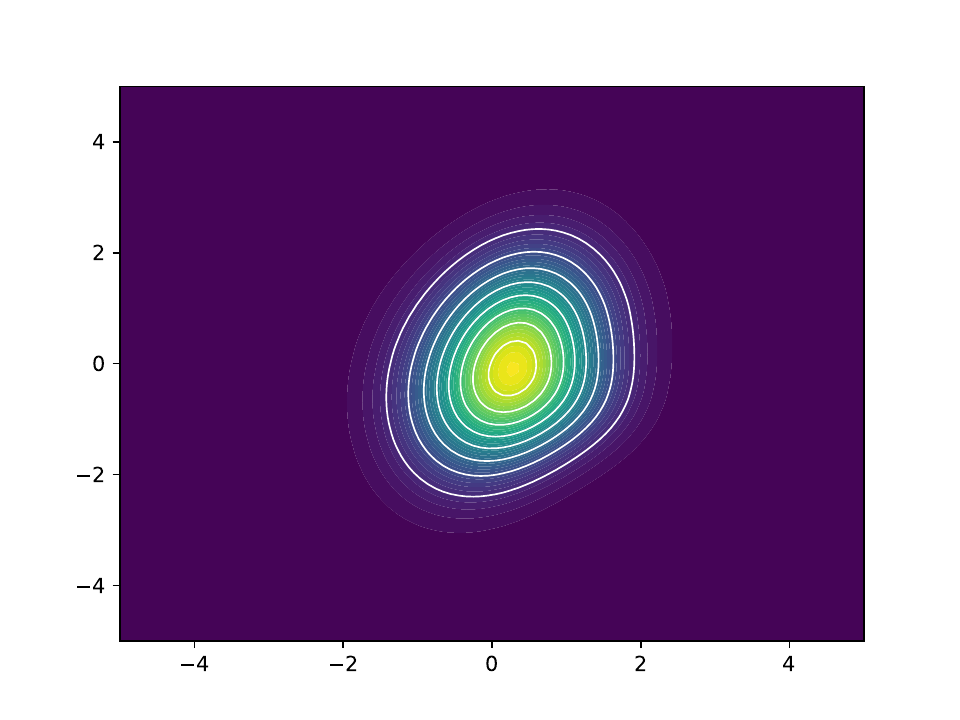} &
        \cropfig{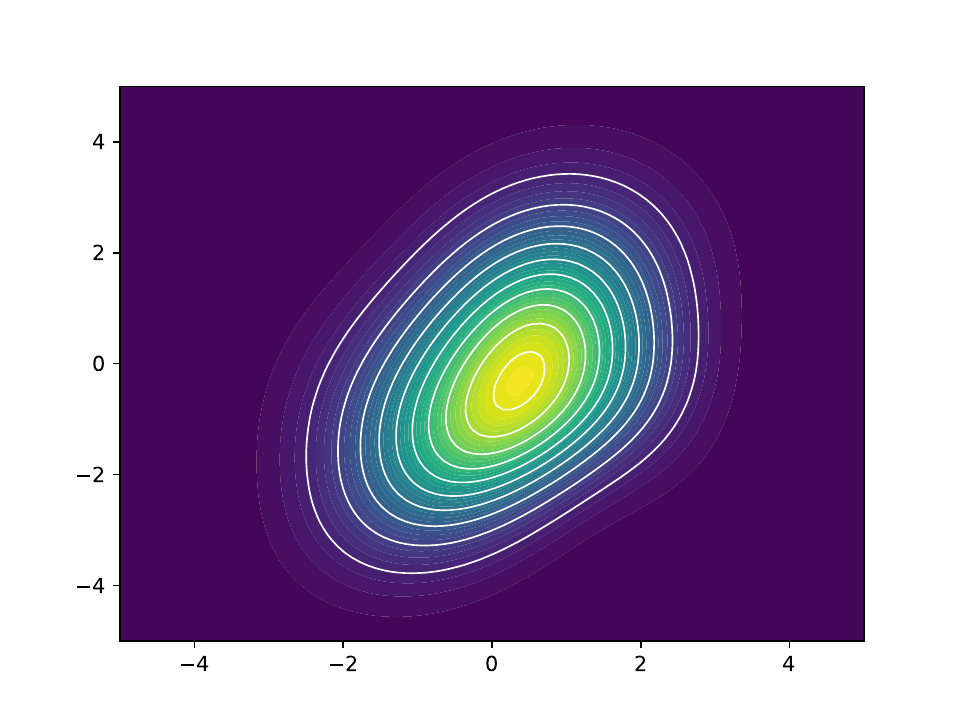} &
        \cropfig{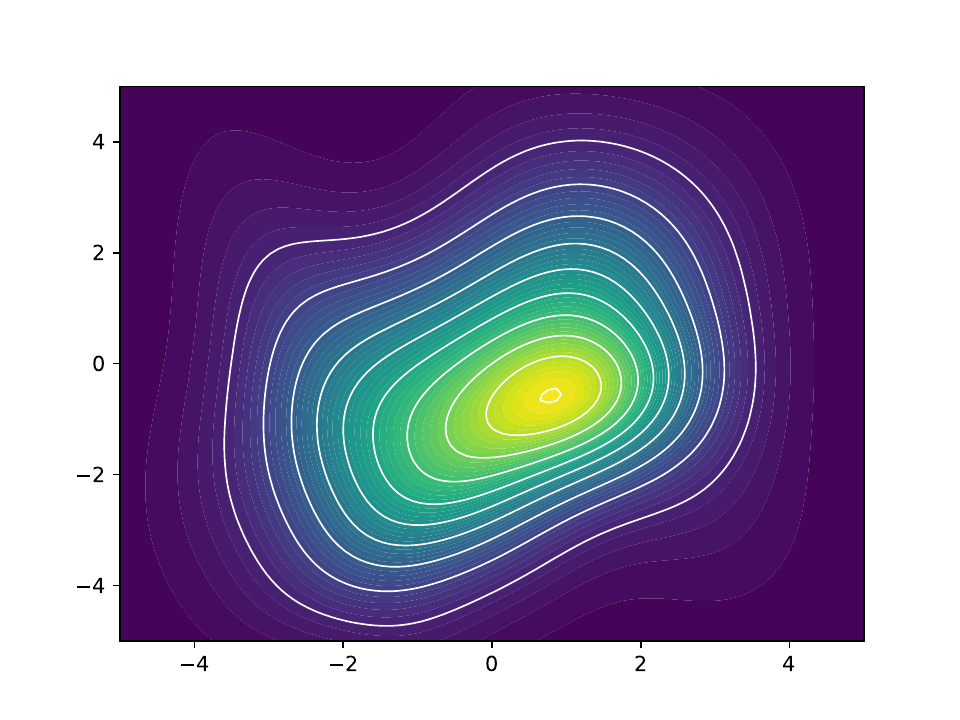} &
        \cropfig{figures/exp_2D_AG/ekf/k_9.pdf} \\
        
        \rotatebox{90}{\strut \scriptsize WSASOS} &
        \cropfig{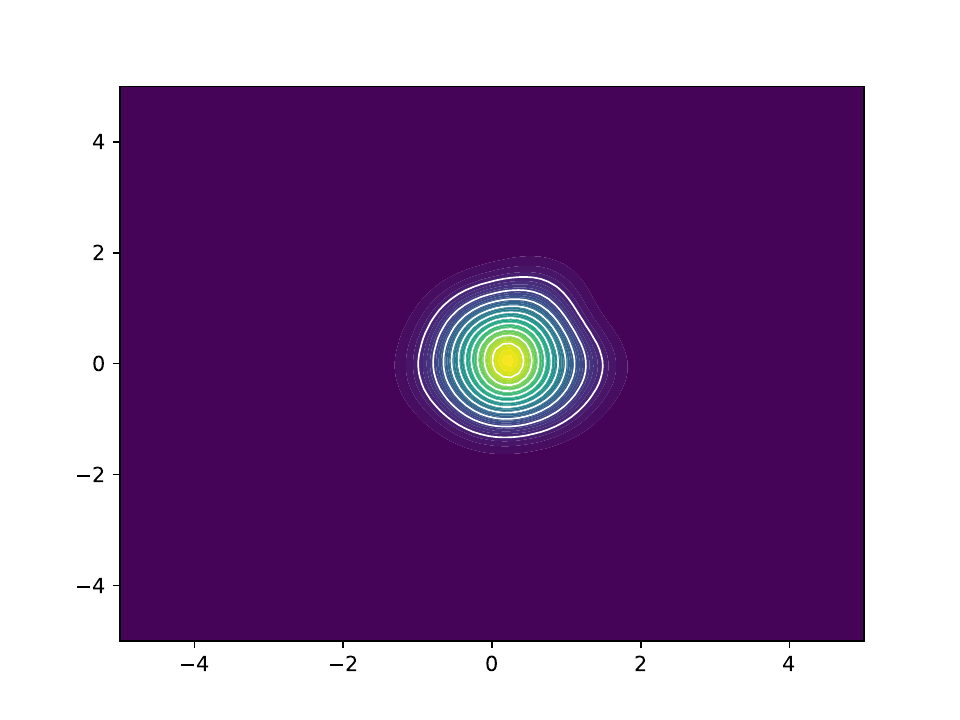} &
        \cropfig{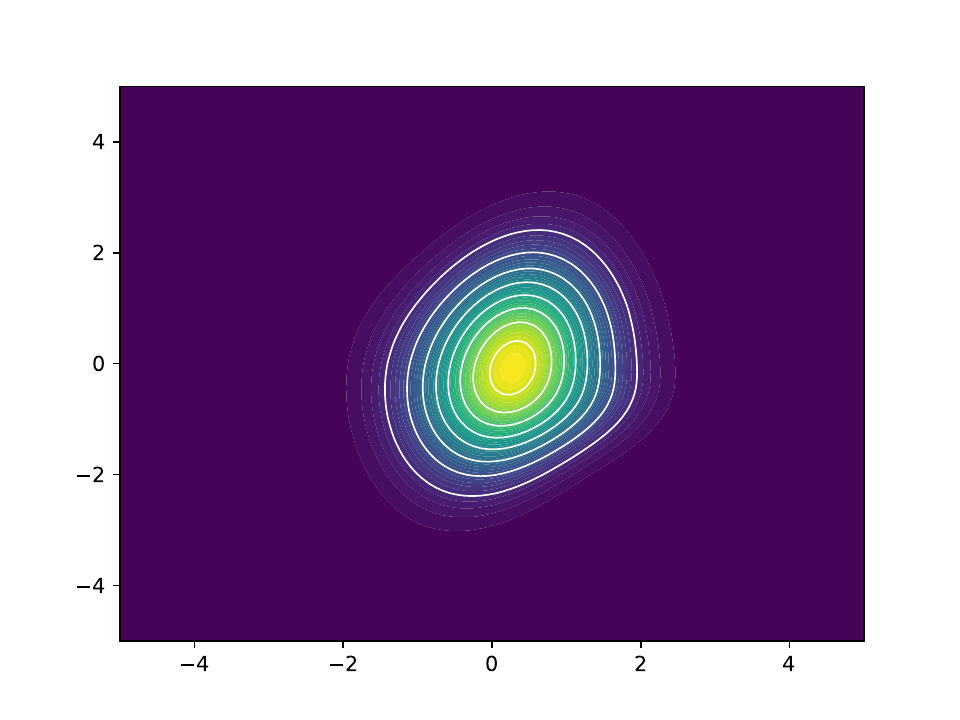} &
        \cropfig{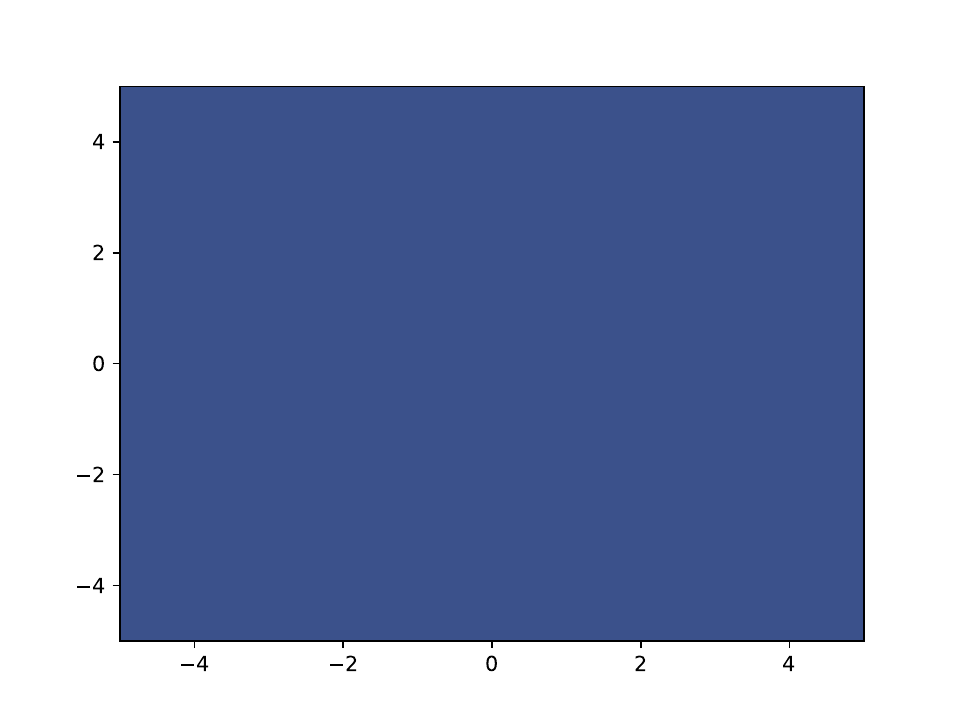} &
        \cropfig{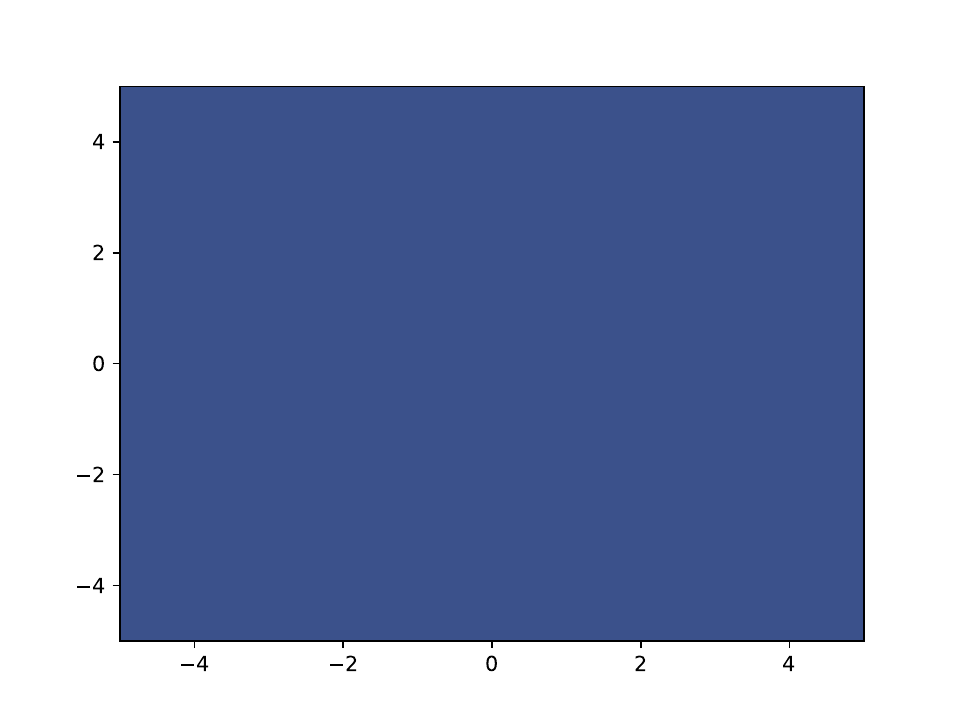} &
        \cropfig{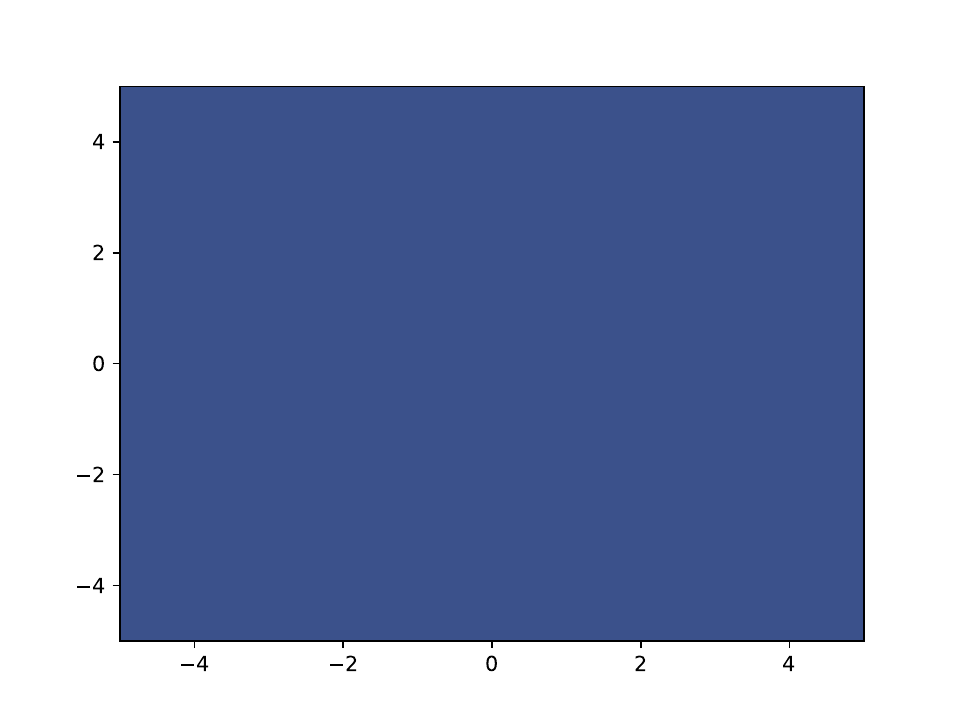} \\
        
        \rotatebox{90}{\strut \scriptsize \textbf{BNF}} &
        \cropfig{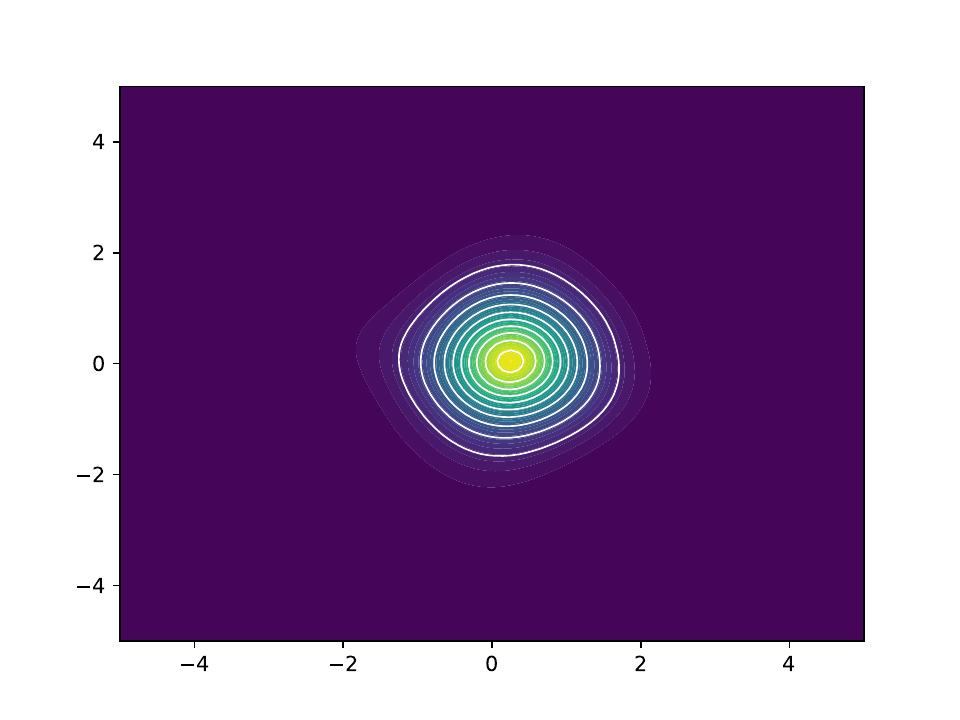} &
        \cropfig{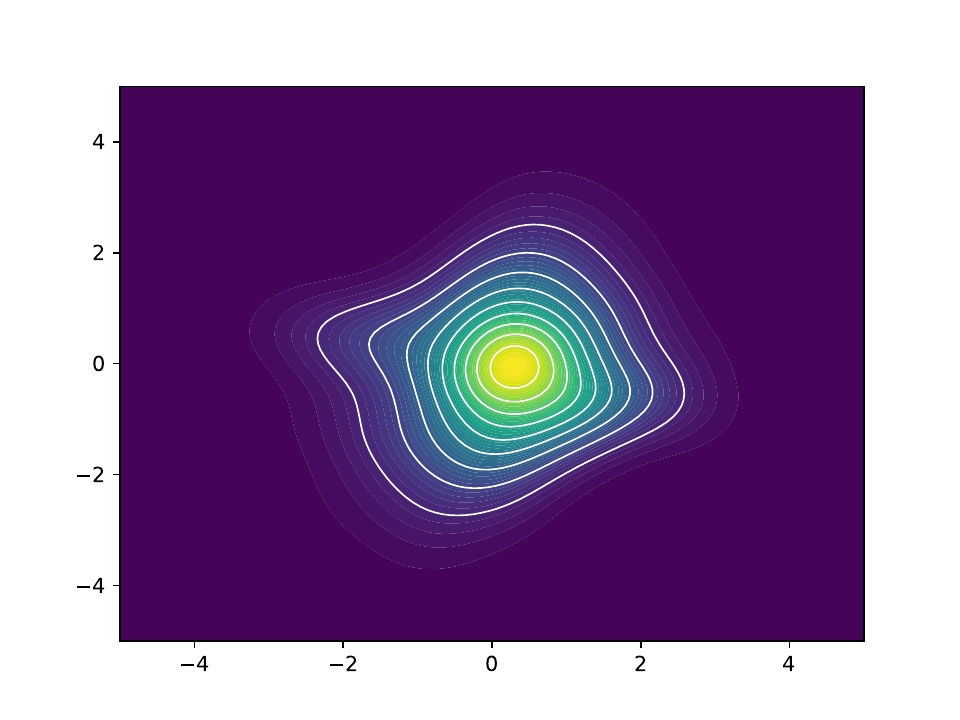} &
        \cropfig{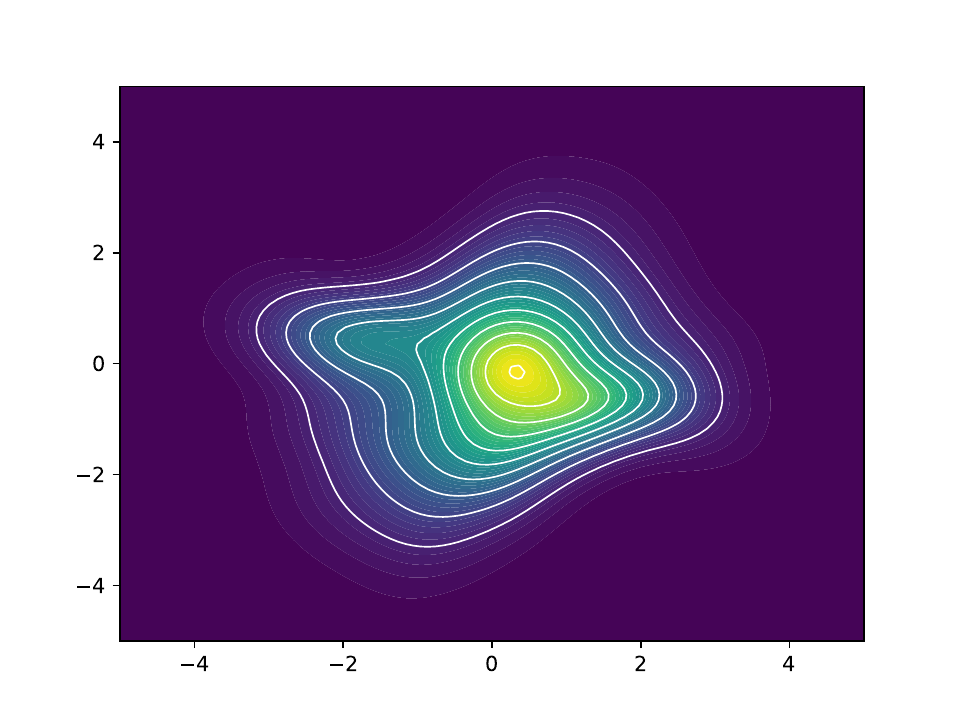} &
        \cropfig{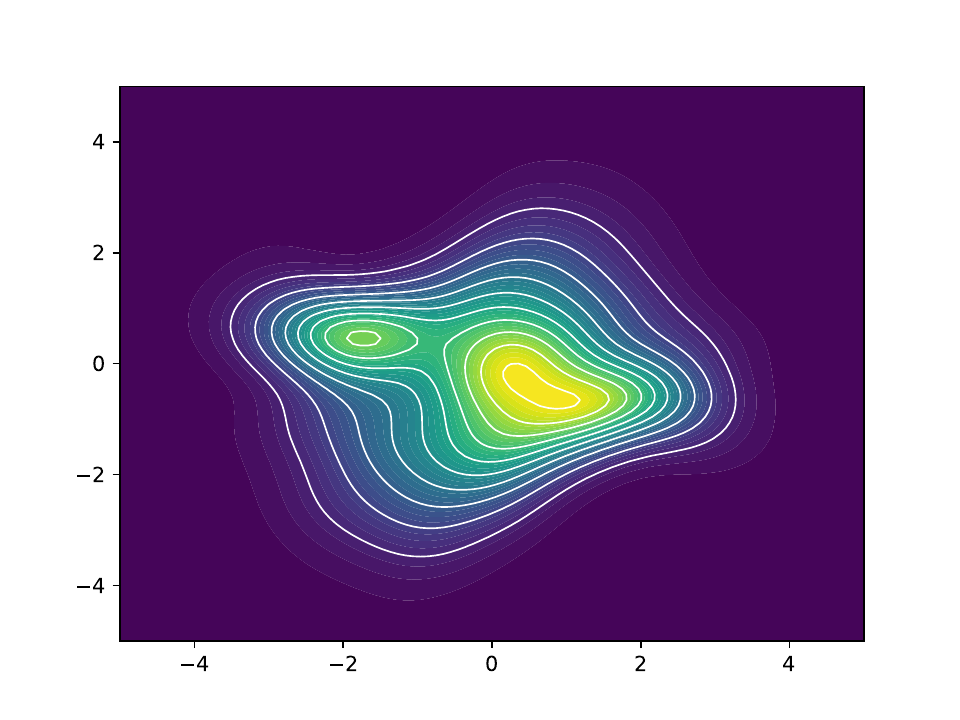} &
        \cropfig{figures/exp_2D_AG/bnf/k_9.pdf} \\
        
        \rotatebox{90}{\strut \scriptsize GridGMM} &
        \cropfig{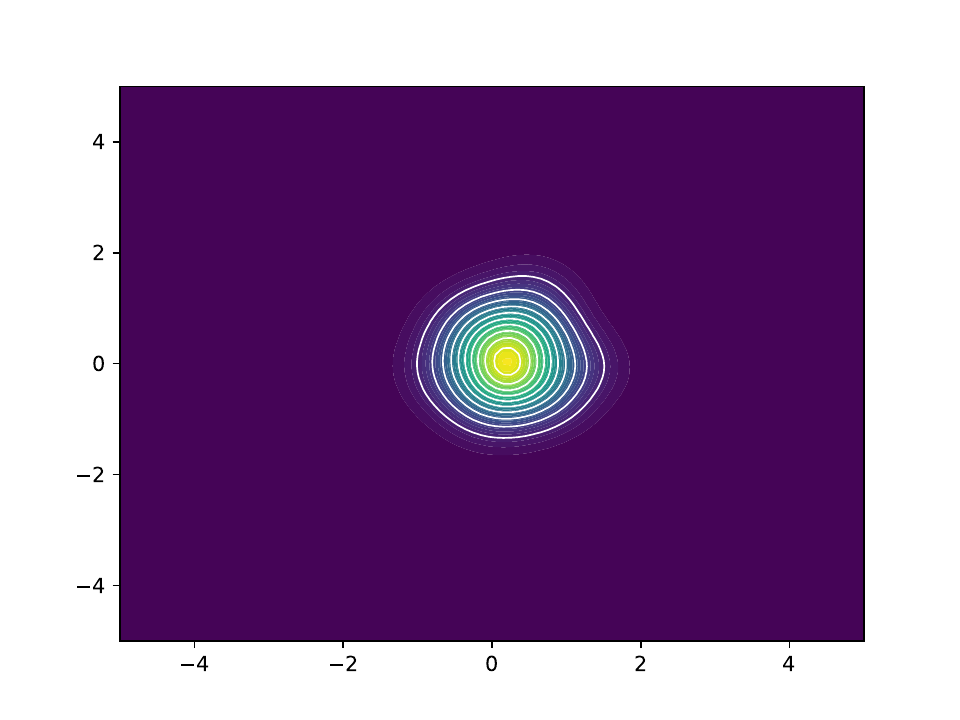} &
        \cropfig{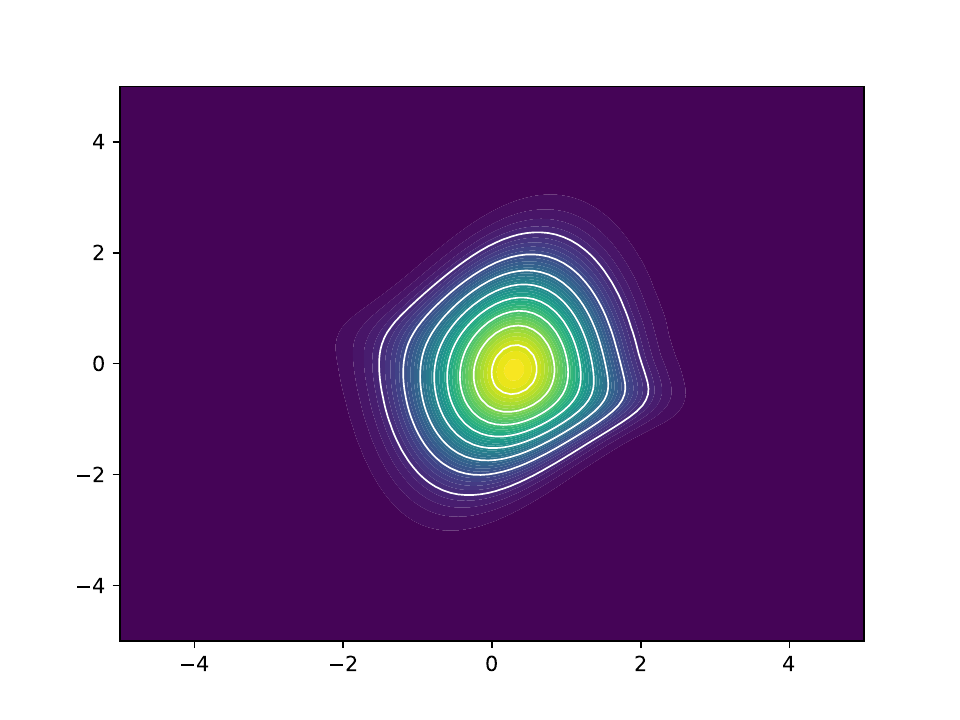} &
        \cropfig{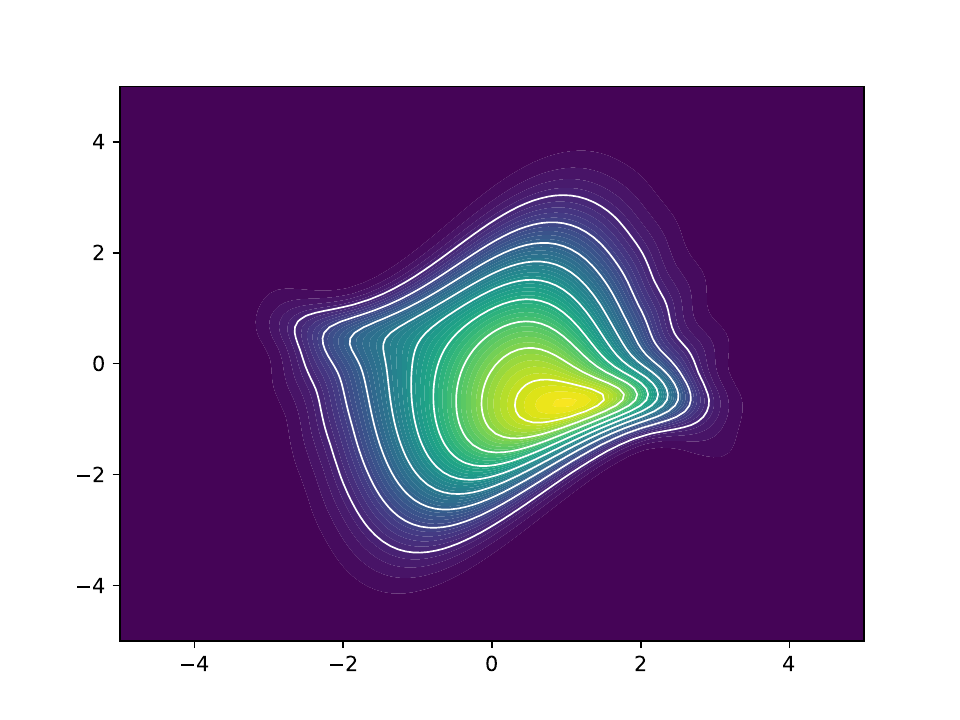} &
        \cropfig{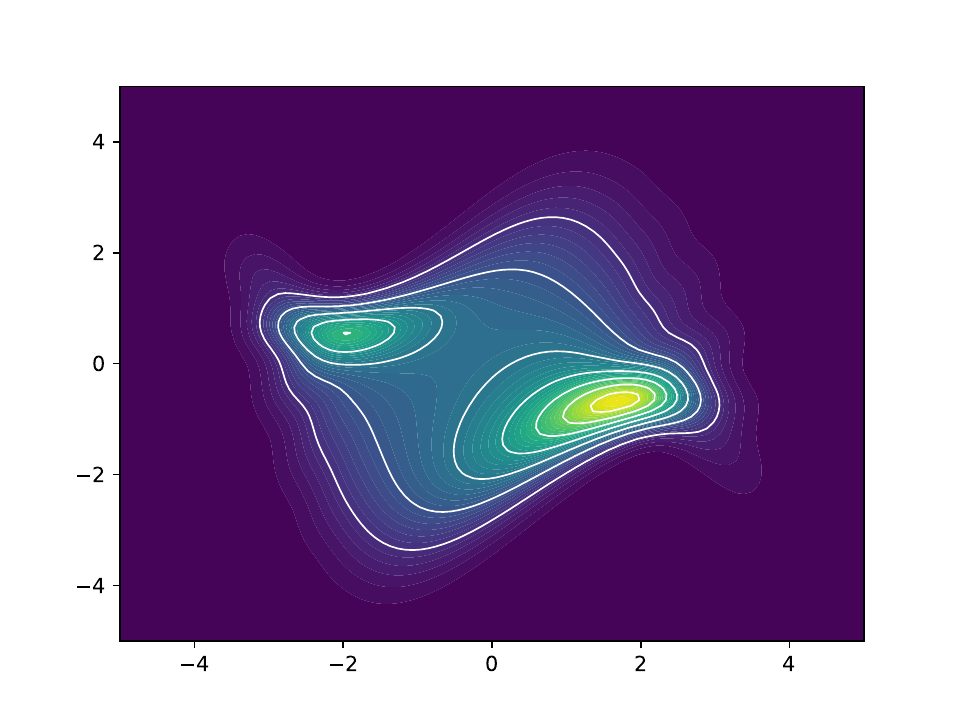} &
        \cropfig{figures/exp_2D_AG/grid_gmm/k_9.pdf} \\
        
        & \footnotesize $k = 1$
        & \footnotesize $k = 3$
        & \footnotesize $k = 5$
        & \footnotesize $k = 7$
        & \footnotesize $k = 9$
    \end{tabular}
    \centering
    \caption{Visual Comparison for non-linear Van der Pol system with additive uncorrelated Gaussian noise}
    \label{fig: visual add gaussian 1-9}
\end{figure*}

\begin{figure*}[t]
    \centering
    \setlength{\tabcolsep}{0pt}
    \begin{tabular}{c*{5}{c}}
        \rotatebox{90}{\strut \scriptsize Monte Carlo} &
        \cropfig{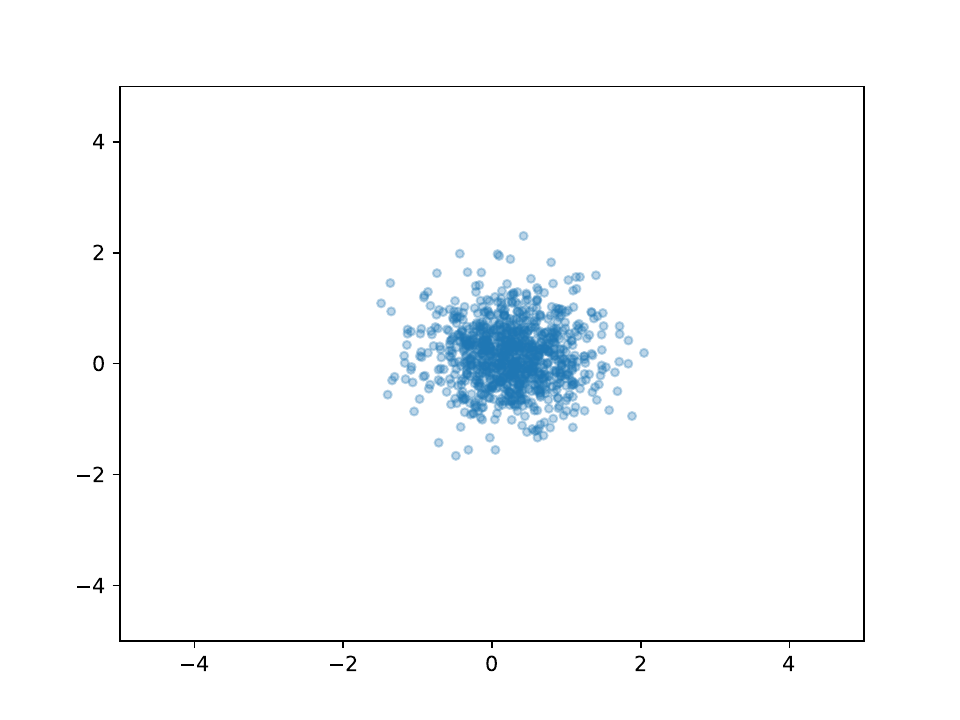} &
        \cropfig{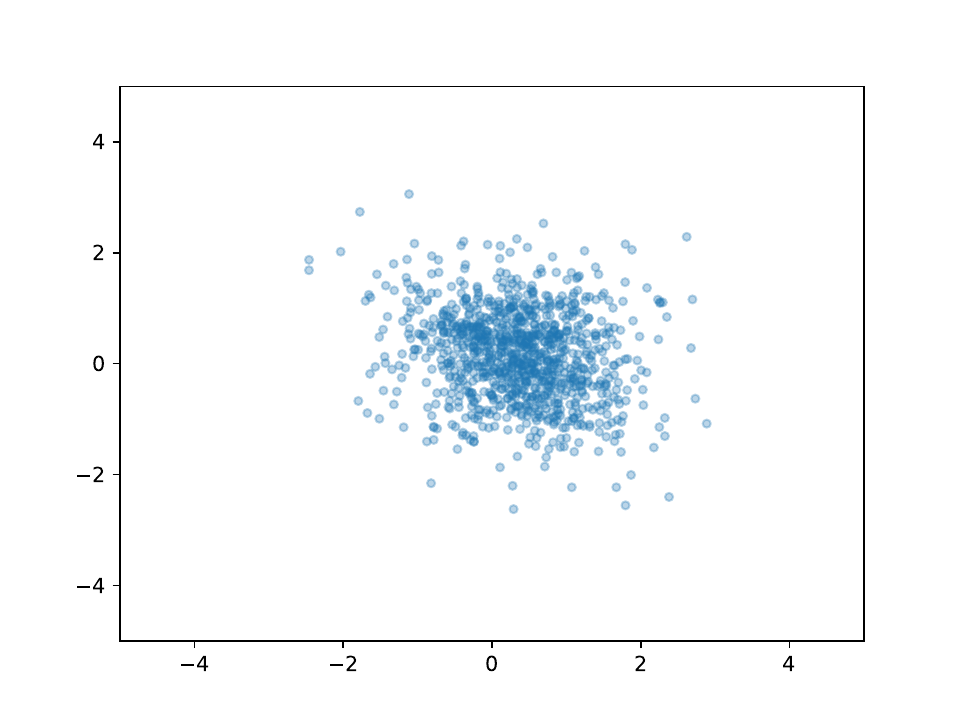} &
        \cropfig{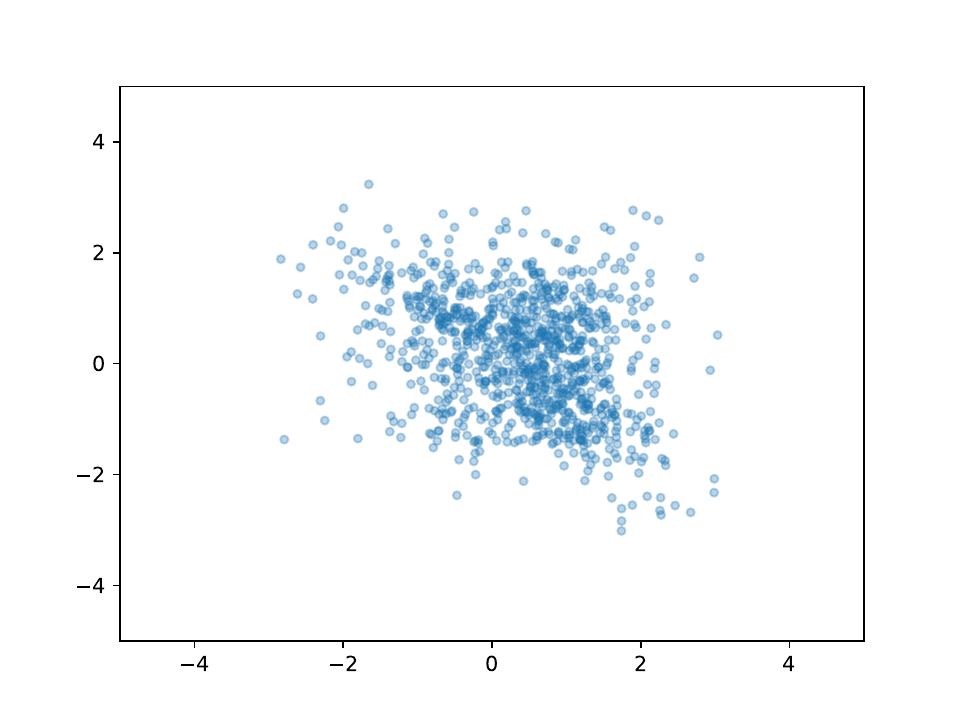} &
        \cropfig{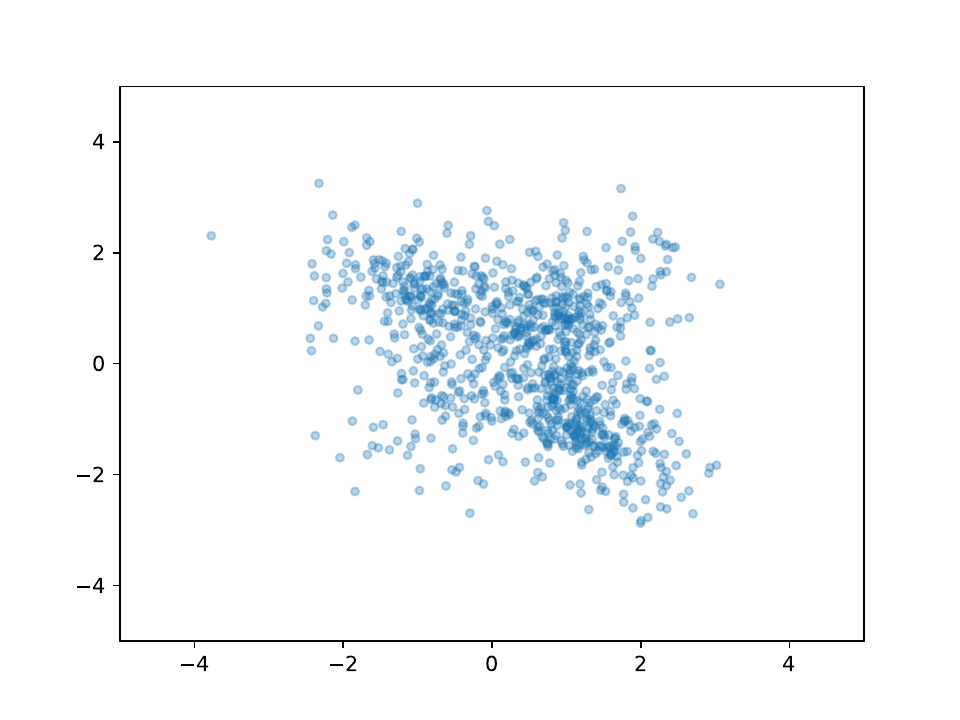} &
        \cropfig{figures/exp_2D_NL/monte_carlo/k_9.pdf} \\

        \rotatebox{90}{\strut \scriptsize EKF} &
        \cropfig{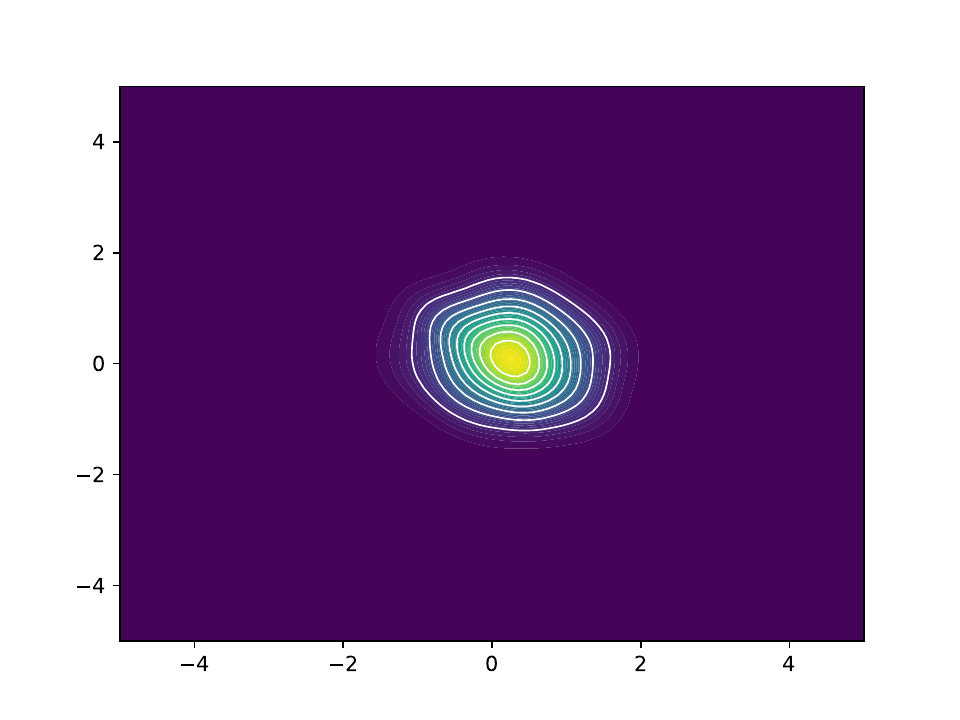} &
        \cropfig{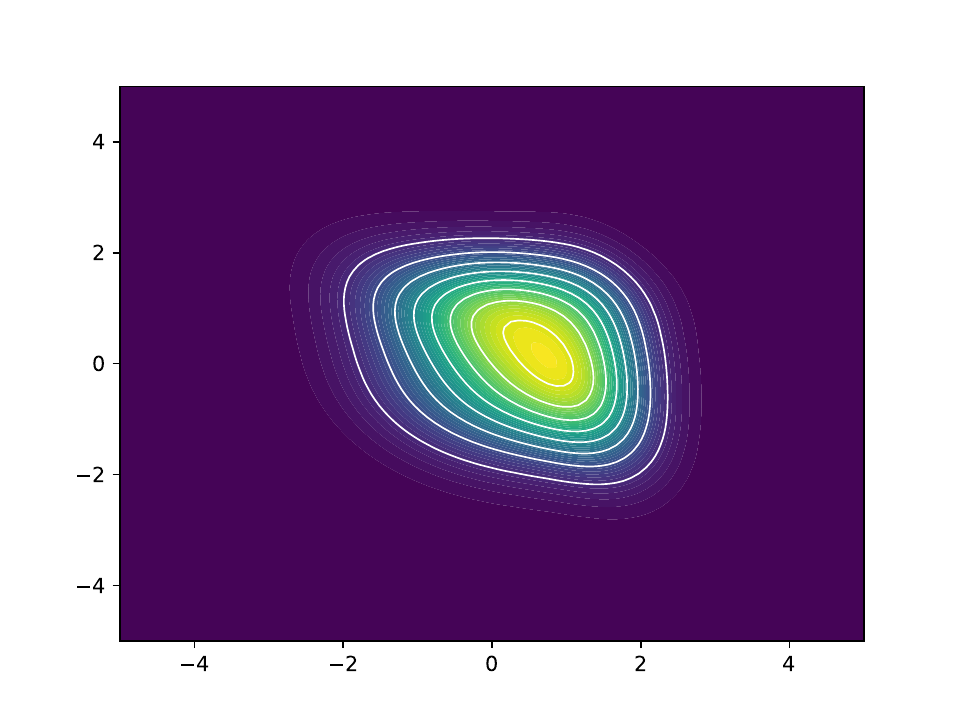} &
        \cropfig{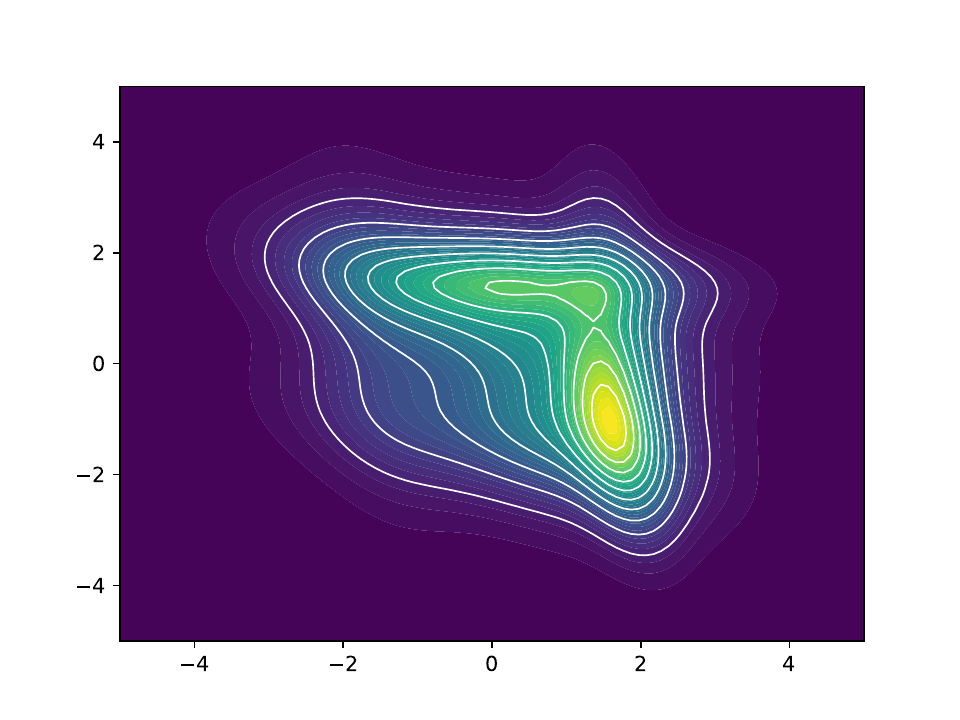} &
        \cropfig{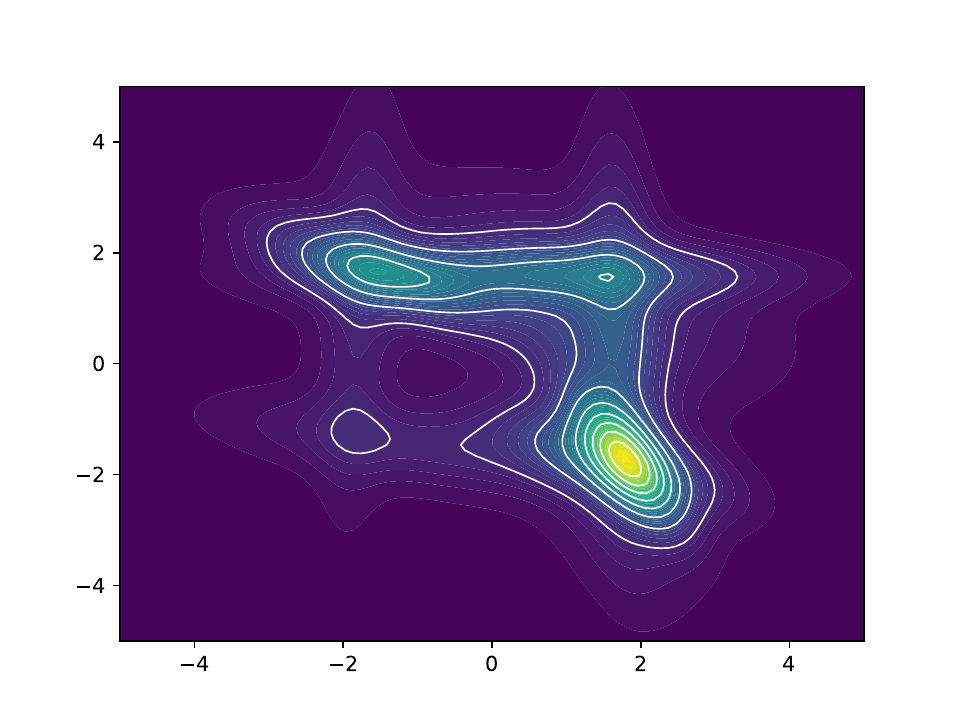} &
        \cropfig{figures/exp_2D_NL/ekf/k_9.pdf} \\
        
        \rotatebox{90}{\strut \scriptsize WSASOS} &
        \cropfig{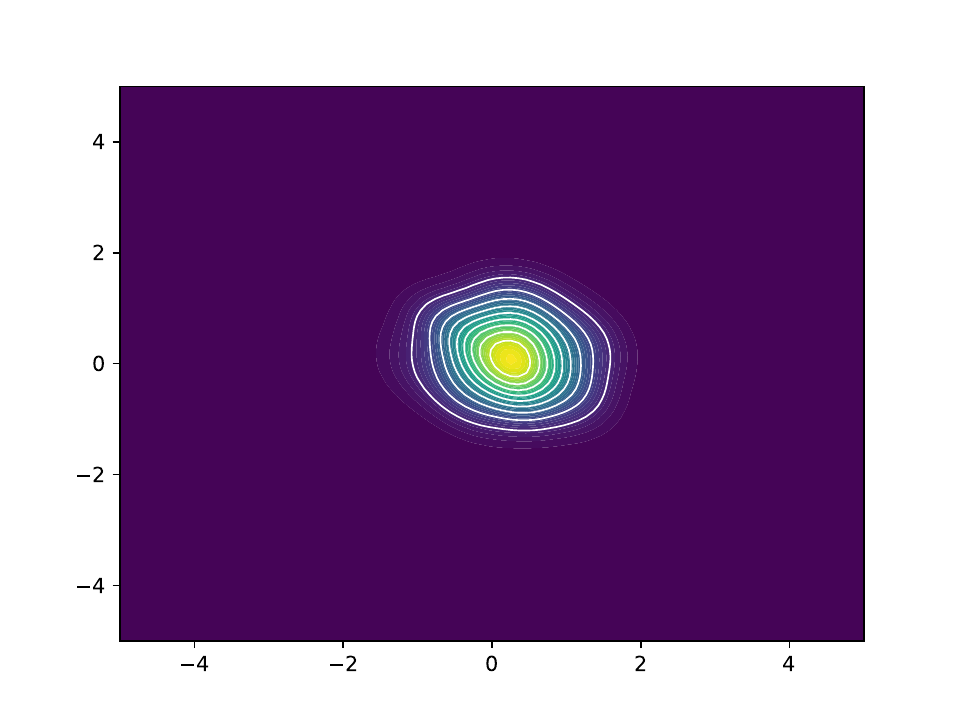} &
        \cropfig{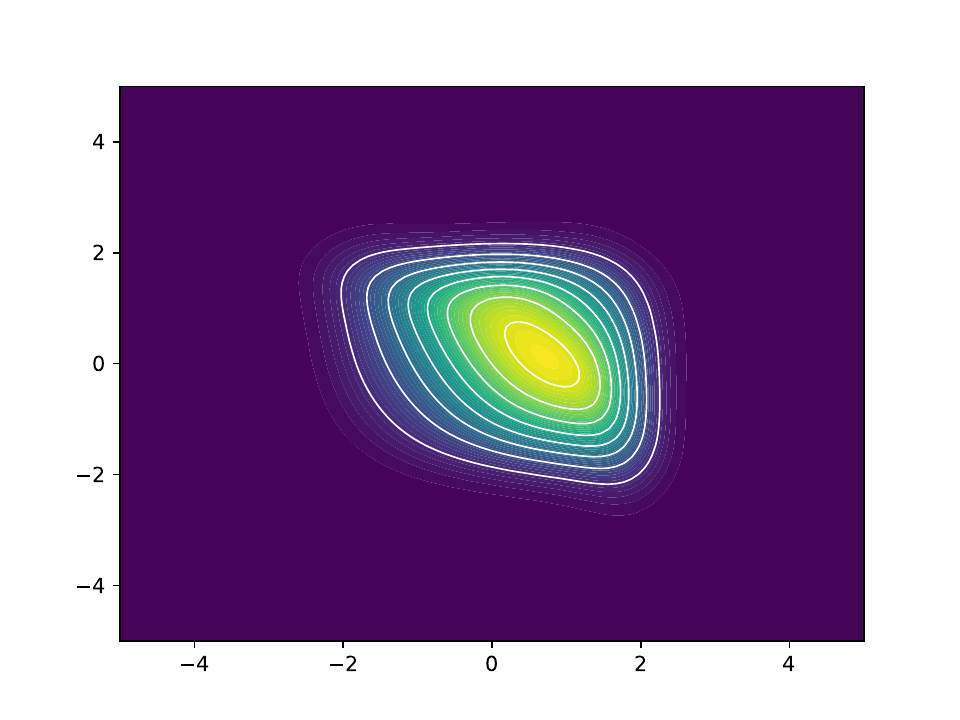} &
        \cropfig{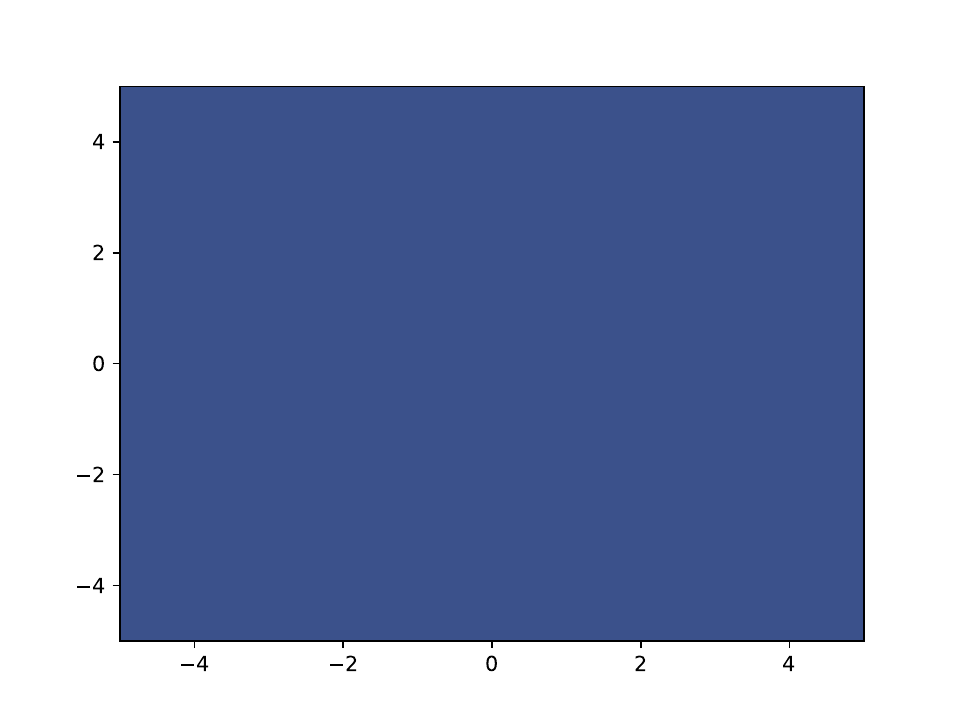} &
        \cropfig{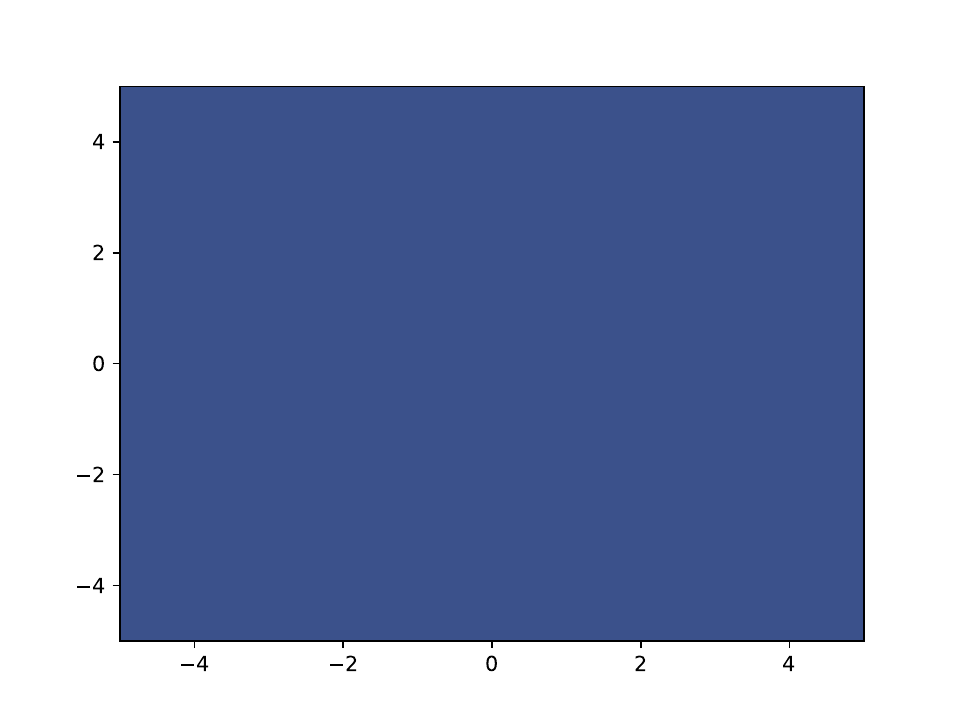} &
        \cropfig{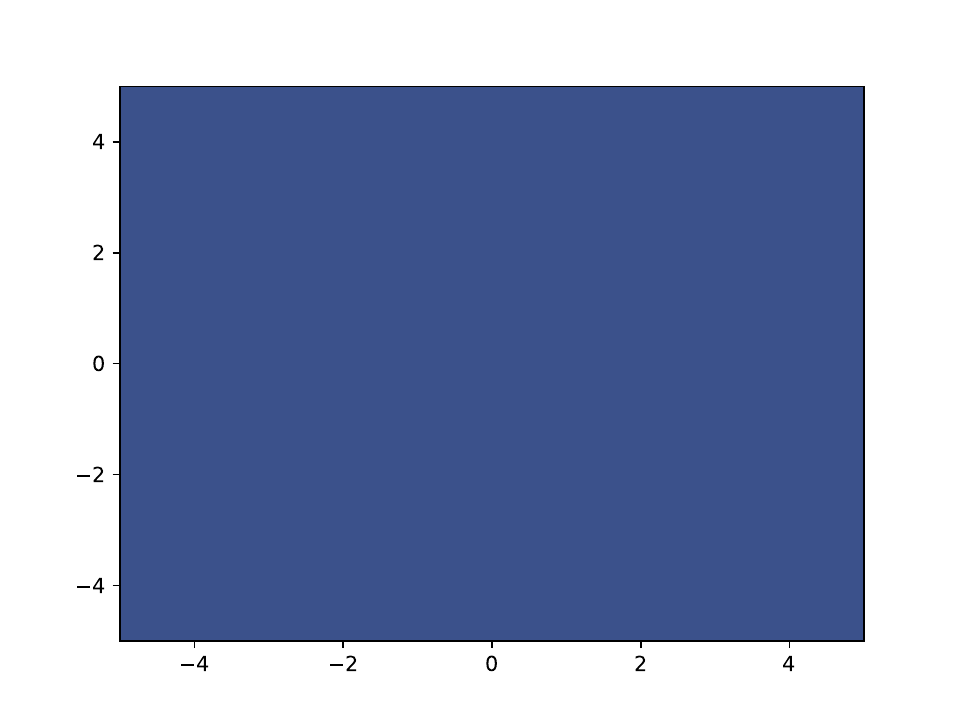} \\
        
        \rotatebox{90}{\strut \scriptsize \textbf{BNF}} &
        \cropfig{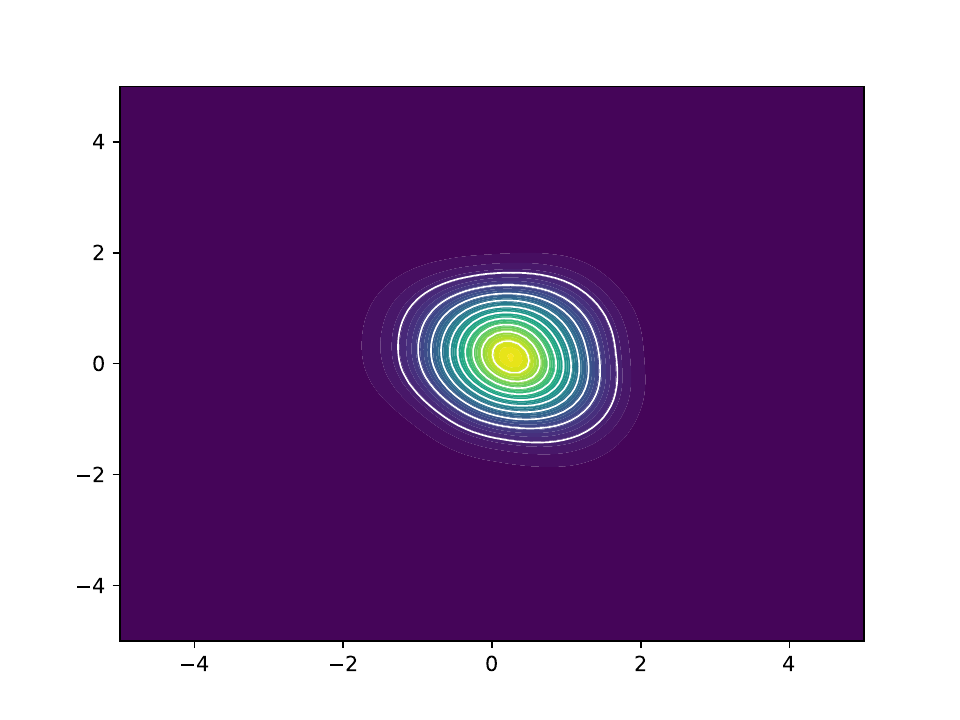} &
        \cropfig{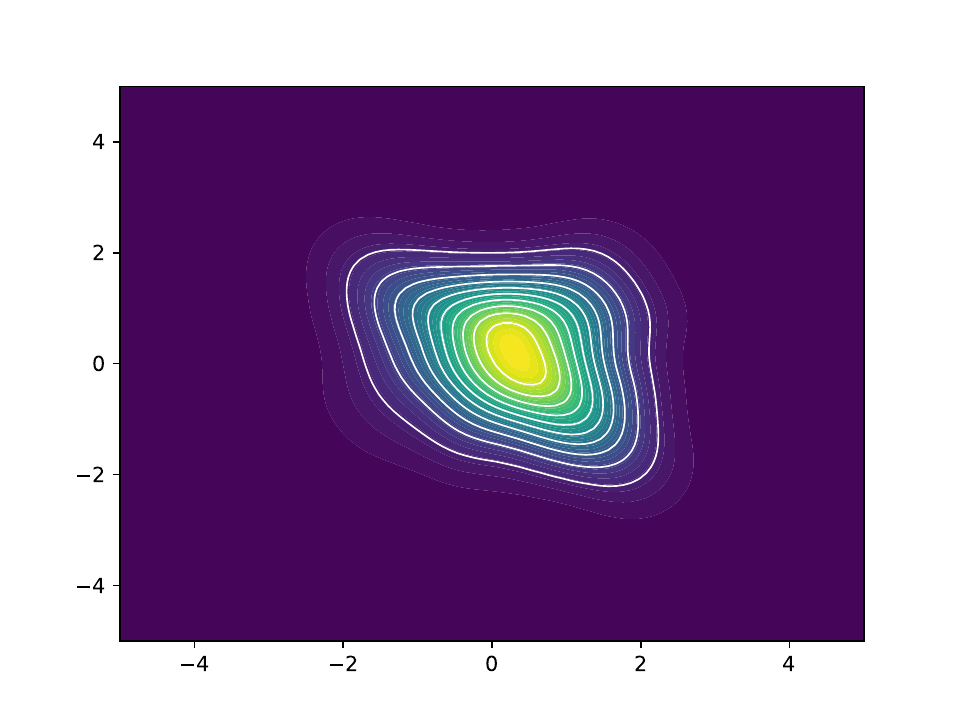} &
        \cropfig{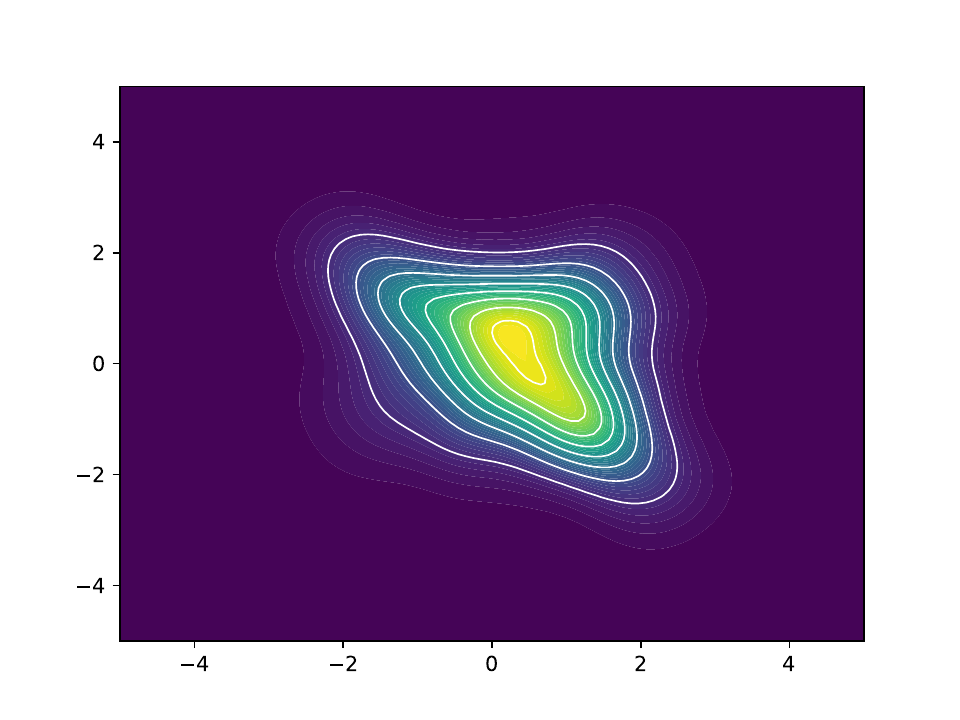} &
        \cropfig{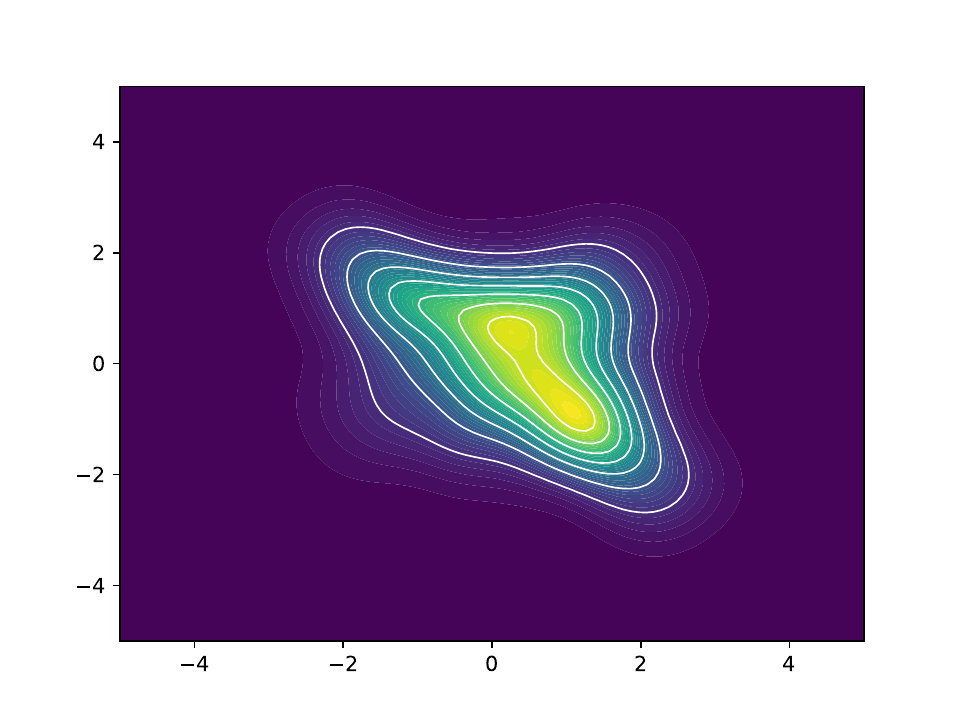} &
        \cropfig{figures/exp_2D_NL/bnf/k_9.pdf} \\
        
        \rotatebox{90}{\strut \scriptsize GridGMM} &
        \cropfig{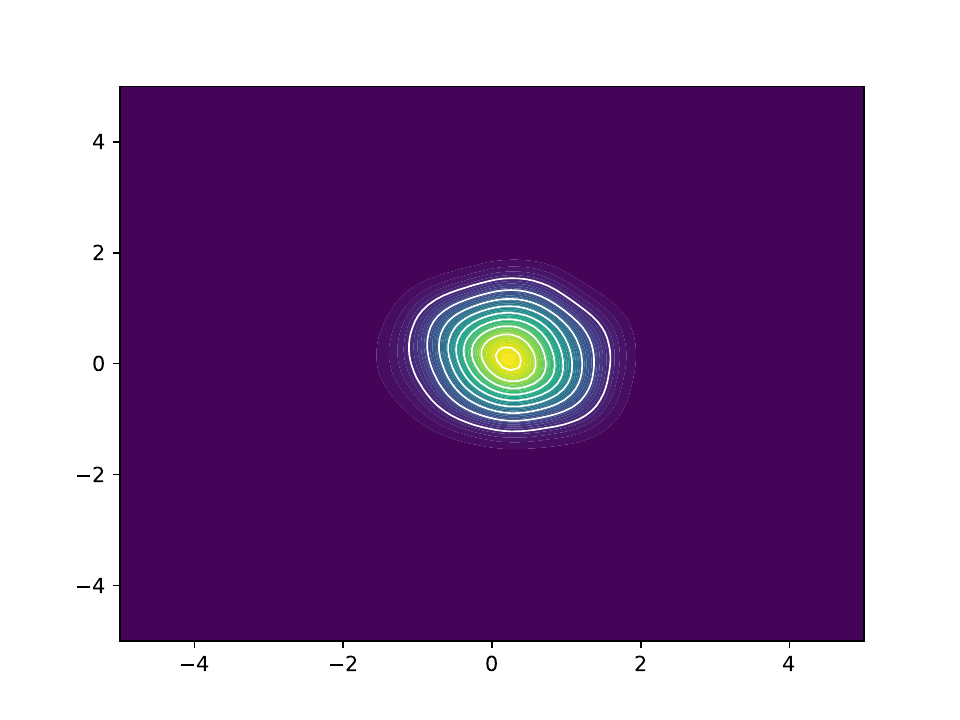} &
        \cropfig{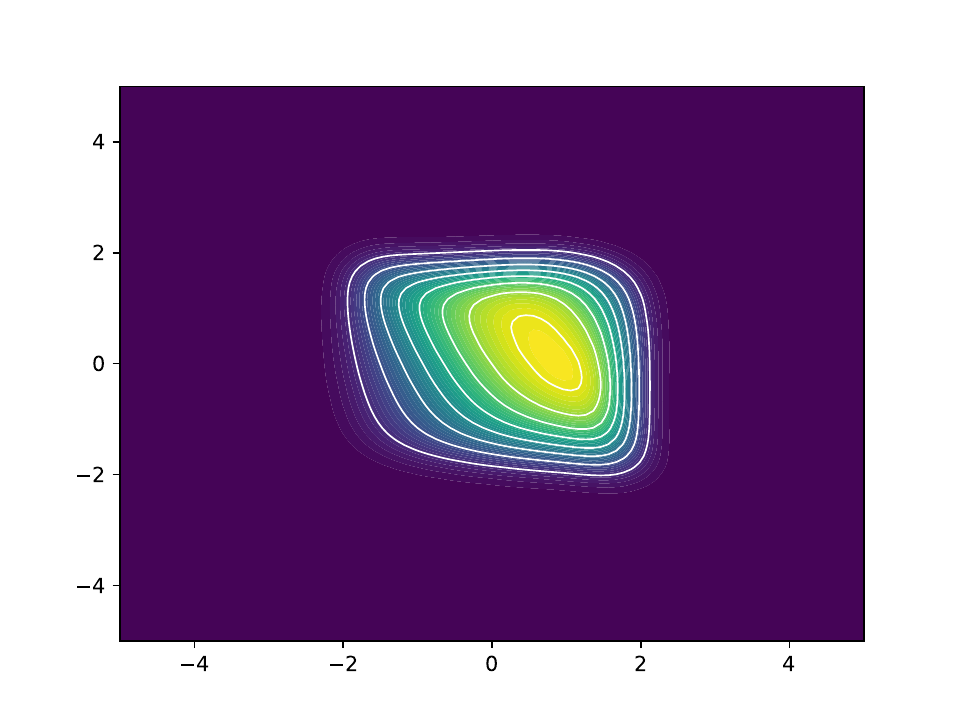} &
        \cropfig{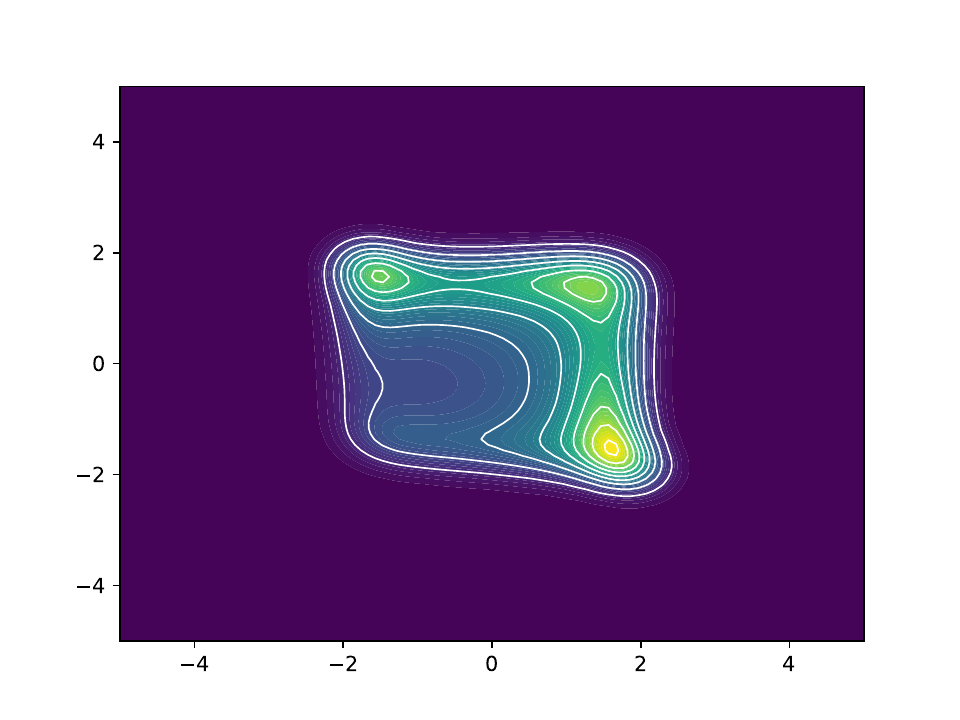} &
        \cropfig{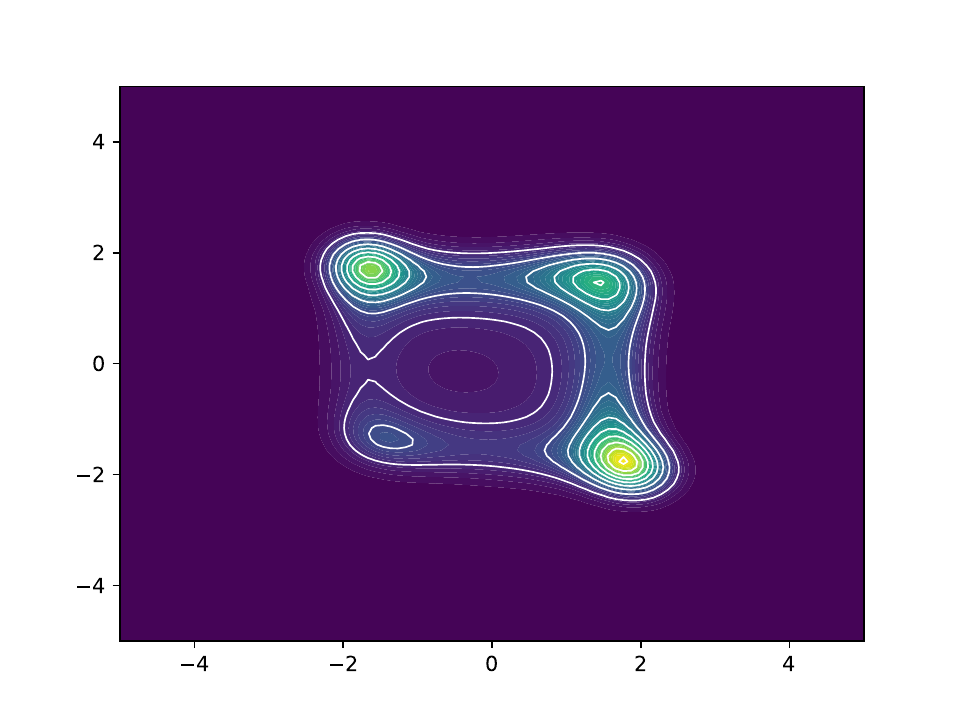} &
        \cropfig{figures/exp_2D_NL/grid_gmm/k_9.pdf} \\
        
        & \footnotesize $k = 1$
        & \footnotesize $k = 3$
        & \footnotesize $k = 5$
        & \footnotesize $k = 7$
        & \footnotesize $k = 9$
    \end{tabular}
    \centering
    \caption{Visual Comparison for non-linear stable attractor system with non-linear multi-modal noise}
    \label{fig: visual non gaussian 1-9}
\end{figure*}


\end{document}